\documentclass{article}

\usepackage{arxiv}

\usepackage[font=small,labelfont=bf]{caption}
\usepackage[utf8]{inputenc} 
\usepackage[T1]{fontenc}    
\usepackage{hyperref}       
\usepackage{url}            
\usepackage{booktabs}       
\usepackage{amsfonts}       
\usepackage{nicefrac}       
\usepackage{microtype}      
\usepackage{lipsum}
\usepackage{graphicx}
\graphicspath{ {./images/} }

\usepackage{pdflscape}   
\usepackage{longtable}   
\usepackage{tabularx}
\newcolumntype{L}[1]{>{\raggedright\arraybackslash}p{#1}}
\newcolumntype{C}[1]{>{\centering\arraybackslash}p{#1}}
\usepackage{makecell}    
\usepackage{pgfplots}
\pgfplotsset{compat=1.18}
\usetikzlibrary{patterns}

\usepackage{tabularx}
\usepackage{siunitx}
\usepackage{array}
\usepackage{makecell}
\usepackage{xcolor}
\usepackage{biblatex}
\definecolor{mygreen}{RGB}{0,128,0}   
\definecolor{myyellow}{RGB}{220,150,0}  

\title{MetaboNet: The Largest Publicly Available Consolidated Dataset for Type 1 Diabetes Management}

\author{
 Miriam K. Wolff \\
  Replica Health\\
  New York, NY, USA \\
  \texttt{miriam@replica.health} \\
   \And
 Peter Calhoun \\
  Jaeb Center for Health Research\\
  Tampa, FL, USA \\
  \texttt{pcalhoun@jaeb.org}
  \And
 Eleonora Maria Aiello \\
  University of Pavia\\
  Pavia, IT \\
  \texttt{eleonoramaria.aiello@unipv.it}
  \And
 Yao Qin \\
  University of Santa Barbara \\
  Santa Barbara, CA, USA \\
  \texttt{yaoqin@ucsb.edu}
  \And
Sam F. Royston \\
  Replica Health \\
  New York, NY, USA \\
  \texttt{sam@replica.health} \\
}

\begin{document}
\maketitle
\fancypagestyle{firstpage}{%
  \fancyhf{}%
  \fancyhead[R]{Accepted at \emph{Journal of Diabetes Science and Technology}, 2026}%
  \fancyfoot[C]{\thepage}%
  \renewcommand{\headrulewidth}{0pt}%
}
\fancyhf{}
\fancyhead[R]{Wolff et al.}
\fancyfoot[C]{\thepage}
\renewcommand{\headrulewidth}{0pt}

\thispagestyle{firstpage}

\begin{abstract}
Progress in Type 1 Diabetes (T1D) algorithm development is limited by the fragmentation and lack of standardization across existing T1D management datasets. Current datasets differ substantially in structure and are time-consuming to access and process, which impedes data integration and reduces the comparability and generalizability of algorithmic developments. This work aims to establish a unified and accessible data resource for T1D algorithm development.
Multiple publicly available T1D datasets were harmonized into a unified resource, termed the MetaboNet dataset. Inclusion required the availability of both continuous glucose monitoring (CGM) data and corresponding insulin pump dosing records. Additionally, auxiliary information such as reported carbohydrate intake and physical activity was retained when present.
The MetaboNet dataset comprises 3135 subjects and 1228 patient-years of overlapping CGM and insulin data, making it substantially larger than existing standalone benchmark datasets. The resource is distributed as a fully public subset available for immediate download at https://metabo-net.org/, and with a Data Use Agreement (DUA)-restricted subset accessible through their respective application processes. For the datasets in the latter subset, processing pipelines are provided to automatically convert the data into the standardized MetaboNet format.
A harmonized public dataset for T1D research is presented, and the access pathways for both its unrestricted and DUA-governed components are described. The resulting dataset covers a broad range of glycemic profiles and demographics and thus can yield more generalizable algorithmic performance than individual datasets.
\end{abstract}

\section{Introduction}
Management of type 1 diabetes (T1D) remains challenging, despite advances in technology and computational methods that have substantially improved patient care \cite{AielloDeshpandeOzaslan2021AIDReview}. Many individuals, regardless of insulin delivery method, fail to achieve glycemic targets or avoid severe hypoglycemic events \cite{LaffelSherrLiu2025LimitationsCGMTargets}. Achieving optimal glycemic control is complicated by the interplay of physiological variability, behavioral and lifestyle factors, such as diet and physical activity.

In diabetes technology research, existing datasets have played a critical role in supporting algorithm development and evaluation \cite{AielloDeshpandeOzaslan2021AIDReview}. The constraints of algorithm development have been increasingly loosened by the growing availability of data, which was previously limited. Building on this foundation, data-driven approaches that leverage continuous glucose monitoring (CGM) and other physiological signals offer the potential to further optimize glycemic control and reduce the risk of hypo- and hyperglycemia. For instance, Kovatchev et al. demonstrated a neural-network-based artificial pancreas achieving a time-in-range (TIR) of 86\% during a 20-hour hotel session \cite{KovatchevFrasquetPryor2024NeuralNetArtificialPancreas}, while Aiello et al. showed that a model predictive controller (MPC) incorporating a data-driven glucose prediction algorithm could outperform traditional linear MPC approaches \cite{AielloJaloliCescon2024MPCArtificialPancreas}. Other examples of data-driven algorithms in T1D management are meal-detection algorithms \cite{TurksoySamadiFeng2016MealDetectionModule} and hypoglycemia prediction models \cite{TurksoyBayrakQuinn2013HypoglycemiaEarlyAlarm}. The development and clinical implementation of these innovations depend on high-quality datasets that represent diverse populations and real-world diabetes management, underscoring the importance of accessible, comprehensive data for advancing research and improving outcomes.

A limitation of existing datasets is that each dataset is collected for a specific purpose, and consequently, these data may represent only a specific aspect of T1D management at the time. In particular, each of these resources provides distinct advantages, such as extended longitudinal coverage \cite{AlsuhaymiBilalGarcia2025LongitudinalMultimodalDataset}, larger cohorts \cite{RiddellLiGal2023T1DEXI}, richer feature sets \cite{MarlingBunescu2020OhioT1DMUpdate}, or varying levels of accessibility, ranging from fully public release to controlled access under data use agreements (DUA). For example, the OhioT1DM dataset has been widely used in T1D research since its introduction in 2018, offering a feature-rich resource that has enabled the exploration of novel hypotheses and algorithm development \cite{MarlingBunescu2020OhioT1DMUpdate}. Despite its value, OhioT1DM is limited in terms of data longevity, demographic representation, and overall size. In response, several newer datasets have been developed, including The Type 1 Diabetes and Exercise Initiative (T1DEXI) \cite{RiddellLiGal2023T1DEXI}, BrisT1D \cite{James2025BrisT1DOpenDataset}, DiaTrend \cite{PrioleauBartolomeComi2023DiaTrend}, T1D-UOM \cite{AlsuhaymiBilalGarcia2025LongitudinalMultimodalDataset}, and AZT1D \cite{KhamesianArefeenThompson2025AZT1D}. These datasets each represent important contributions to the field, yet when used individually, they may still face limitations in supporting comprehensive, standardized research and benchmarking.

These resources vary in size, access procedures, and data formatting, which can create additional preprocessing work for researchers and limit the ease of cross-study comparisons. In Maheshwari et al. \cite{MaheshwariKaliaTewari2025AIForDiabetesReview}, the authors state that artificial intelligence models often exhibit bias due to non-representative datasets, limiting their generalizability across diverse populations. Cinar et al. addressed this gap by introducing DiaData \cite{cinar2025benchmarking,cinar2025diadata}, integrating 15 datasets of 2510 subjects. DiaData primarily centers on CGM data, demographics and heart rate data. While this presents a substantial contribution to the field, key determinants to glucose variability, such as insulin administration, meal intake, and physical activity, are not incorporated. 

With this aim in mind, we introduce MetaboNet, a harmonized, multi-source dataset for T1D research, provided in a standardized tabular format with uniform sampling time, to facilitate use in any machine-learning applications. Inclusion required the availability of both continuous glucose monitoring (CGM) data and corresponding insulin pump dosing records. Additionally, auxiliary information such as meal intake, physical activity, device data, and demographics was retained when present. 

MetaboNet is named after and inspired by how ImageNet \cite{DengDongSocher2009ImageNet} transformed advances in deep learning within computer vision \cite{ChawlaNakovAli2023TenYearsAfterImageNet}. It consolidates multiple existing datasets into a unified structure, facilitating cross-study analyses in T1D research, with an aim of improving benchmark-driven research in metabolic research. A substantial portion of the dataset is publicly available, while additional data are available under their respective DUA; for these, we provide standardized processing pipelines to ensure consistency with the public release.

\section{Methods}
\label{sec:headings}
\begin{figure}
    \centering
    \includegraphics[width=1\linewidth]{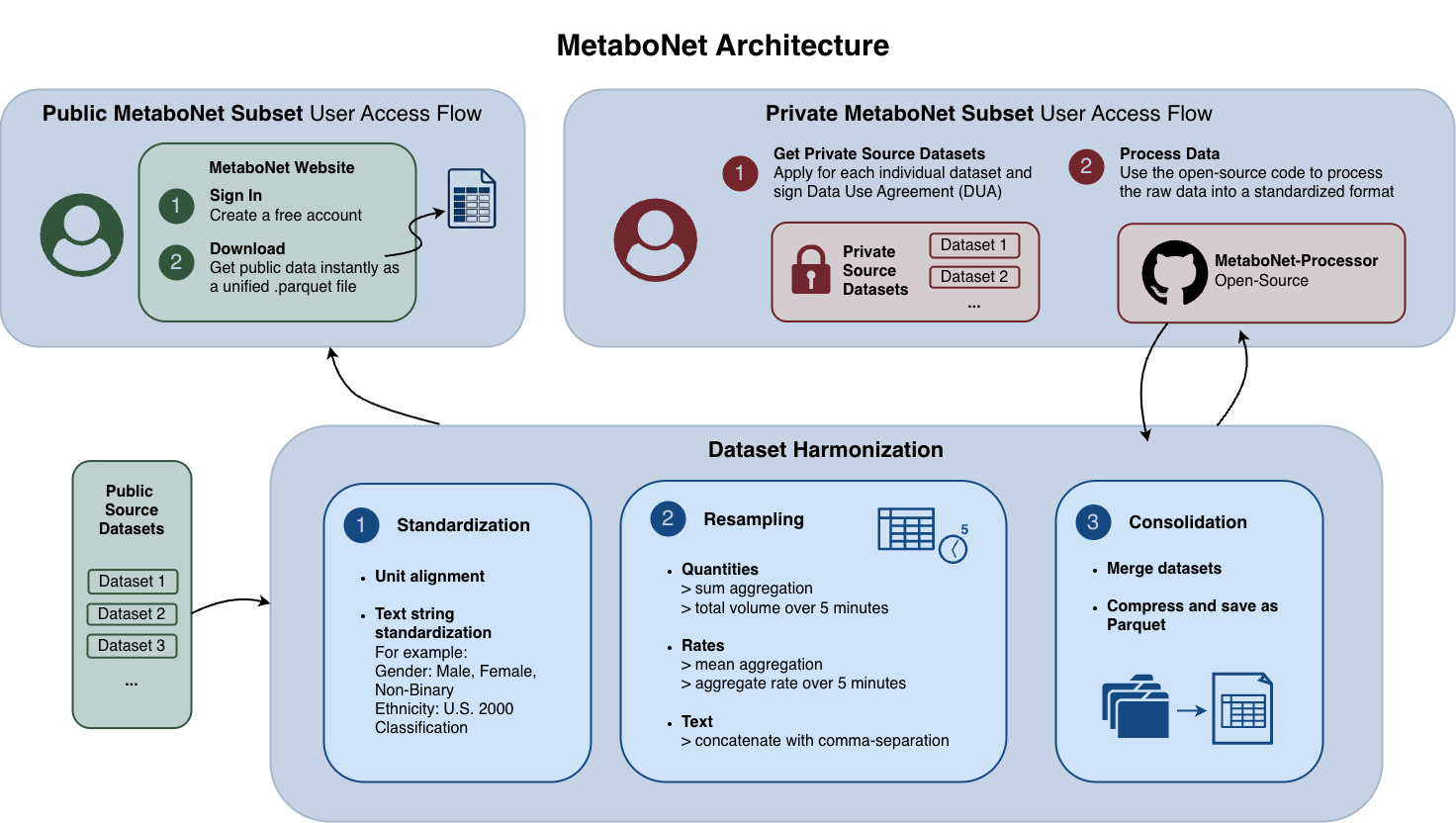}
    \caption{Figure illustrating the dataset access process for the public and private subsets of Metabonet, as well as the dataset harmonization pipeline. }
    \label{fig:fig_1_v2}
\end{figure}

For this study, we harmonized multiple publicly available T1D management datasets to create a unified resource, as shown in the pipeline in Figure \ref{fig:fig_1_v2}. Inclusion criteria required that datasets provide both continuous glucose monitoring (CGM) and insulin dosing information, while encompassing data from both multiple daily injection (MDI) and insulin pump users. We prioritized datasets with permissive licensing, larger cohorts, and extended longitudinal coverage, capturing data from each subject over longer observational windows. Datasets meeting these criteria were identified through a targeted search in the JAEB dataset repository \cite{AnonymousNDDPublicStudyWebsitesJAEB}, the Babelbetes \cite{AnonymousNDDBabelbetes} project, the review of open dynamic glycemic data in diabetes research by Del Giudice et al. \cite{DelGiudicePiersantiGobl2025OpenDynamicGlycemicData}, and other publicly available datasets meeting the same requirements \cite{AlsuhaymiBilalGarcia2025LongitudinalMultimodalDataset,James2025BrisT1DOpenDataset,PrioleauBartolomeComi2023DiaTrend,ShahidLewis2022LargeScaleDataAnalysisGlucoseVariability,NeinsteinWongLook2016TidepoolPlatform}. We welcome the community to reach out regarding any additional datasets for incorporation into future releases.

\subsection{MetaboNet Public Subset}

Table \ref{tab:1} provides an overview of the datasets included in MetaboNet 2026. Under \textit{Availability}, datasets included in the public subset are marked in green. Some of these are distributed under the permissive license Creative Commons Attribution 4.0 International. In addition to these publicly licensed datasets, MetaboNet has established a partnership and obtained formal approval from JAEB to integrate and utilize their data. Consequently, these datasets can be processed, consolidated, and redistributed within MetaboNet in compliance with their licensing terms.

\subsection{MetaboNet DUA-Restricted Subset}

In Table 1, DUA-Restricted datasets within MetaboNet are highlighted with green availability. For this subset, MetaboNet does not have redistribution rights to the dataset, and users must access the individual source datasets through their respective application processes. We provide a processing script that allows researchers to locally convert DUA-governed datasets to the MetaboNet-compatible format in an open-source code repository \cite{ReplicahealthNDMetaboNetGitHub}.

\sisetup{
  detect-weight=true,
}
\newgeometry{top=0.2cm, bottom=1.2cm, left=1.2cm, right=1.2cm}

\begin{landscape}
\begin{table}[p]
\parbox{0.92\linewidth}{\caption{Overview of the included datasets in MetaboNet 2026. The numbers are reported after processing. The ``Monitoring Period'' is defined according to the study-reported primary outcome measures and may differ slightly from the actual data. ``Years of data'' is derived from the total rows, based on a 5-minute sampling interval. Datasets marked in \textcolor{mygreen}{green} indicate that MetaboNet has redistribution rights; datasets marked in \textcolor{myyellow}{yellow} do not. Diabetes duration, weight, and age are reported as mean~$\pm$~std followed by median~[IQR].}
\label{tab:1}
}
\centering
\tiny
\setlength{\tabcolsep}{2pt}
\renewcommand{\arraystretch}{1.25}
\begin{tabular}{%
  L{3.0cm}   
  C{0.75cm}  
  L{2.0cm}   
  C{2.5cm}   
  L{1.5cm}  
  C{2.25cm}  
  C{2.1cm}   
  L{2.9cm}   
  C{1.3cm}   
  C{0.9cm}   
  C{1.0cm}   
  L{2.5cm}   
}
\hline
\textbf{Dataset Name} &
\textbf{N} &
\textbf{Study Design} &
\textbf{Age [years]} &
\textbf{Sex} &
\textbf{Weight [lbs]} &
\textbf{T1D Duration [years]} &
\textbf{CGM Device} &
\textbf{Monitoring Period} &
\textbf{Total yrs} &
\textbf{CGM+Ins yrs} &
\textbf{Availability} \\
\hline
\multicolumn{12}{l}{\textit{\textbf{MetaboNet Public Subset}}} \\
\hline
CTR3~\cite{jaeb_ctr3_581} &
30 &
Randomized, open-label, crossover &
\makecell{40.3$\pm$12.8 \\ 44.0 [29.5--50.5]} &
\makecell[l]{Male: 56.7\% \\ Female: 43.3\%} &
\makecell{168.5$\pm$29.7 \\ 169.9 [146.9--189.2]} &
\makecell{21.6$\pm$11.2 \\ 19.0 [13.8--28.8]} &
Dexcom G4: 100\% &
2 weeks &
13.5 &
4.3 &
\textcolor{mygreen}{Public by JAEB DUA} \\
 
DCLP3~\cite{jaeb_idcl_dclp3_573} &
112 &
RCT &
\makecell{32.7$\pm$15.7 \\ 28.5 [17.0--44.2]} &
\makecell[l]{Male: 51.8\% \\ Female: 48.2\%} &
\makecell{169.7$\pm$37.1 \\ 162.6 [144.2--190.3]} &
\makecell{19.5$\pm$14.4 \\ 16.5 [7.8--28.0]} &
Dexcom G6: 100\% &
26 weeks &
57.4 &
48.5 &
\textcolor{mygreen}{Public by JAEB DUA} \\
 
DCLP5~\cite{jaeb_dclp5_535} &
100 &
RCT &
\makecell{10.6$\pm$2.0 \\ 11.0 [9.0--12.0]} &
\makecell[l]{Male: 51.0\% \\ Female: 49.0\%} &
\makecell{95.1$\pm$30.9 \\ 89.0 [71.4--106.6]} &
\makecell{5.2$\pm$2.8 \\ 5.0 [3.0--7.0]} &
Dexcom G6: 100\% &
16 weeks &
58.0 &
45.9 &
\textcolor{mygreen}{Public by JAEB DUA} \\
 
Flair~\cite{jaeb_flair_566} &
113 &
Randomized, open-label, crossover &
\makecell{19.3$\pm$4.2 \\ 19.0 [16.0--22.0]} &
\makecell[l]{Female: 61.9\% \\ Male: 38.1\%} &
\makecell{161.7$\pm$32.4 \\ 160.5 [134.0--181.2]} &
\makecell{11.3$\pm$5.5 \\ 12.0 [7.0--15.0]} &
Guardian 3: 100\% &
12 weeks &
66.6 &
38.9 &
\textcolor{mygreen}{Public by JAEB DUA} \\
 
IOBP2~\cite{jaeb_ilet_579_iobp2} &
332 &
RCT &
\makecell{31.8$\pm$19.2 \\ 30.5 [14.0--48.0]} &
\makecell[l]{Female: 50.9\% \\ Male: 49.1\%} &
\makecell{162.3$\pm$51.5 \\ 164.9 [128.4--196.9]} &
\makecell{18.0$\pm$14.3 \\ 14.0 [6.0--28.0]} &
Dexcom G6: 100\% &
13 weeks &
92.1 &
54.5 &
\textcolor{mygreen}{Public by JAEB DUA} \\
 
Loop Observational Study~\cite{jaeb_loop_560} &
845 &
Prospective observational cohort &
\makecell{27.2$\pm$17.2 \\ 27.0 [11.0--39.0]} &
\makecell[l]{Female: 51.6\% \\ Male: 41.2\% \\ Missing: 7.1\% \\ Non-Binary: 0.1\%} &
\makecell{136.1$\pm$55.6 \\ 141.0 [95.0--173.8]} &
\makecell{15.0$\pm$13.0 \\ 11.0 [4.0--23.0]} &
\makecell[l]{G6: 75.5\%, G5: 15.1\% \\ Missing: 7.2\% \\ G4: 1.8\%, Libre: 0.2\% \\ Enlite: 0.1\%} &
12 months &
901.9 &
519.0 &
\textcolor{mygreen}{Public by JAEB DUA} \\
 
PEDAP~\cite{jaeb_pedap_599} &
65 &
RCT &
\makecell{3.3$\pm$1.2 \\ 4.0 [2.0--4.0]} &
\makecell[l]{Male: 52.3\% \\ Female: 47.7\%} &
\makecell{38.8$\pm$10.2 \\ 37.0 [32.6--43.0]} &
\makecell{1.4$\pm$0.9 \\ 1.0 [1.0--2.0]} &
Dexcom G6: 100\% &
13 weeks &
52.4 &
23.2 &
\textcolor{mygreen}{Public by JAEB DUA} \\
 
ReplaceBG~\cite{jaeb_cgm_546_replacebg} &
208 &
RCT &
\makecell{44.1$\pm$13.7 \\ 43.0 [31.0--55.0]} &
\makecell[l]{Female: 48.6\% \\ Male: 48.1\% \\ F: 1.9\%, M: 1.4\%} &
\makecell{178.5$\pm$34.3 \\ 174.7 [152.6--200.6]} &
\makecell{23.4$\pm$11.9 \\ 22.0 [15.0--31.0]} &
Dexcom G4: 100\% &
6 months &
198.5 &
123.6 &
\textcolor{mygreen}{Public by JAEB DUA} \\
 
AZT1D~\cite{KhamesianArefeenThompson2025AZT1D} &
23 &
Real-world observational &
\makecell{58.3$\pm$15.5 \\ 65.0 [46.0--68.0]} &
\makecell[l]{Male: 52.2\% \\ Female: 47.8\%} &
N/A &
N/A &
Dexcom G6: 100\% &
6--8 weeks &
2.7 &
1.7 &
\textcolor{mygreen}{CC BY 4.0} \\
 
BrisT1D~\cite{James2025BrisT1DOpenDataset} &
19 &
Real-world observational &
\makecell{22.1$\pm$1.8 \\ 22.0 [21.0--23.5]} &
\makecell[l]{Female: 73.7\% \\ Male: 15.8\% \\ Non-Binary: 10.5\%} &
N/A &
\makecell{12.7$\pm$4.9 \\ 13.0 [10.0--15.5]} &
\makecell[l]{G6: 52.6\% \\ Libre 2: 26.3\% \\ Guardian 4: 15.8\% \\ G ONE: 5.3\%} &
6 months &
10.1 &
5.8 &
\textcolor{mygreen}{CC BY 4.0} \\
 
HUPA-UCM~\cite{HidalgoAlvaradoBotella2024HUPAUCM} &
22 &
Real-world observational &
\makecell{39.6$\pm$12.8 \\ 41.5 [27.3--48.4]} &
\makecell[l]{Male: 54.5\% \\ Female: 45.5\%} &
\makecell{155.8$\pm$32.3 \\ 148.4 [131.2--175.0]} &
\makecell{17.3$\pm$11.0 \\ 14.9 [10.7--23.1]} &
FreeStyle Libre 2: 100\% &
$\geq$14 days &
2.9 &
1.8 &
\textcolor{mygreen}{CC BY 4.0} \\
 
Shanghai T1DM~\cite{ZhaoZhuShen2023ChineseDiabetesDatasets} &
12 &
Real-world observational &
\makecell{57.8$\pm$11.1 \\ 58.5 [52.5--66.2]} &
\makecell[l]{Female: 58.3\% \\ Male: 41.7\%} &
\makecell{126.6$\pm$22.8 \\ 132.3 [112.9--141.6]} &
\makecell{9.9$\pm$8.4 \\ 8.5 [2.8--16.2]} &
N/A &
3--14 days &
0.4 &
0.1 &
\textcolor{mygreen}{CC BY 4.0} \\
 
T1D-UOM~\cite{AlsuhaymiBilalGarcia2025LongitudinalMultimodalDataset} &
14 &
Real-world observational &
\makecell{46.6$\pm$15.2 \\ 50.0 [31.5--59.0]} &
\makecell[l]{Female: 50.0\% \\ Male: 50.0\%} &
\makecell{174.6$\pm$42.2 \\ 167.6 [139.4--191.3]} &
N/A &
N/A &
3 months &
7.7 &
0.1 &
\textcolor{mygreen}{CC BY 4.0} \\
 
\hline
\textbf{Total Public} &
\textbf{1895} &
{--} &
\makecell{\textbf{29.0$\pm$18.4} \\ \textbf{27.0 [13.0--42.0]}} &
\makecell[l]{\textbf{Female: 51.5\%} \\ \textbf{Male: 45.2\%} \\ \textbf{Missing: 3.2\%} \\ \textbf{Non-Binary: 0.2\%}} &
\makecell{\textbf{144.8$\pm$55.4} \\ \textbf{150.0 [112.0--181.8]}} &
\makecell{\textbf{15.6$\pm$13.1} \\ \textbf{12.0 [5.0--24.0]}} &
\makecell[l]{\textbf{G6: 67.5\%, G4: 13.0\%} \\ \textbf{G5: 6.8\%, Guard.3: 6.0\%} \\ \textbf{Missing: 5.0\%, Libre 2: 1.4\%} \\ \textbf{Guard.4: 0.2\%, others $<$0.2\%}} &
{--} &
\textbf{1464.2} &
\textbf{867.4} &
{--} \\
\hline
\multicolumn{12}{l}{\textit{\textbf{MetaboNet DUA-Governed Subset}}} \\
\hline
DiaTrend~\cite{PrioleauBartolomeComi2023DiaTrend} &
17 &
Real-world observational &
\makecell{51.9$\pm$10.9 \\ 59.5 [39.5--59.5]} &
\makecell[l]{Female: 58.8\% \\ Male: 41.2\%} &
N/A &
N/A &
\makecell[l]{Guardian: 47.1\% \\ G6: 35.3\%, G5: 11.8\% \\ Guardian 3: 5.9\%} &
Varying &
4.0 &
2.6 &
\textcolor{myyellow}{DUA-restricted} \\
 
OhioT1DM~\cite{MarlingBunescu2020OhioT1DMUpdate} &
12 &
Real-world observational &
N/A &
\makecell[l]{Male: 58.3\% \\ Female: 41.7\%} &
N/A &
N/A &
Medtronic Enlite: 100\% &
8 weeks &
1.8 &
1.5 &
\textcolor{myyellow}{DUA-restricted} \\
 
OpenAPS Commons &
173 &
Real-world observational &
\makecell{35.8$\pm$15.1 \\ 35.3 [25.5--45.3]} &
\makecell[l]{Missing: 56.6\% \\ Male: 29.5\% \\ Female: 13.9\%} &
\makecell{163.6$\pm$44.1 \\ 165.3 [140.0--186.0]} &
\makecell{21.6$\pm$14.8 \\ 21.0 [9.5--29.9]} &
N/A &
Varying &
191.1 &
88.0 &
\textcolor{myyellow}{DUA-restricted} \\
 
T1DEXI~\cite{RiddellLiGal2023T1DEXI} &
493 &
Randomized exercise intervention &
\makecell{36.7$\pm$14.0 \\ 33.0 [25.0--46.0]} &
\makecell[l]{Female: 73.0\% \\ Male: 27.0\%} &
\makecell{161.6$\pm$30.1 \\ 156.5 [140.0--180.0]} &
N/A &
Dexcom G6: 100\% &
4 weeks &
55.3 &
31.1 &
\textcolor{myyellow}{DUA-restricted} \\
 
T1DEXIP &
245 &
Real-world observational &
\makecell{13.8$\pm$1.5 \\ 14.0 [12.0--15.0]} &
\makecell[l]{Male: 58.0\% \\ Female: 42.0\%} &
\makecell{129.1$\pm$31.1 \\ 125.0 [108.2--150.0]} &
N/A &
Dexcom G6: 100\% &
10 days &
27.0 &
5.8 &
\textcolor{myyellow}{DUA-restricted} \\
 
Tidepool HCL150~\cite{NeinsteinWongLook2016TidepoolPlatform} &
150 &
Real-world observational &
\makecell{29.1$\pm$16.2 \\ 30.0 [13.0--40.8]} &
\makecell[l]{Female: 44.0\% \\ Male: 31.3\% \\ Missing: 24.7\%} &
N/A &
\makecell{15.8$\pm$12.1 \\ 13.5 [5.2--23.8]} &
N/A &
Varying &
83.3 &
56.8 &
\textcolor{myyellow}{DUA-restricted} \\
 
Tidepool PA50~\cite{NeinsteinWongLook2016TidepoolPlatform} &
50 &
Real-world observational &
\makecell{38.5$\pm$13.1 \\ 37.0 [29.0--49.8]} &
\makecell[l]{Male: 46.0\% \\ Female: 42.0\% \\ Missing: 12.0\%} &
N/A &
\makecell{21.7$\pm$13.0 \\ 20.5 [12.0--30.8]} &
N/A &
Varying &
30.0 &
22.3 &
\textcolor{myyellow}{DUA-restricted} \\
 
Tidepool SAP100~\cite{NeinsteinWongLook2016TidepoolPlatform} &
100 &
Real-world observational &
\makecell{35.9$\pm$20.7 \\ 36.0 [14.0--53.0]} &
\makecell[l]{Missing: 64.0\% \\ Male: 23.0\% \\ Female: 13.0\%} &
N/A &
\makecell{19.8$\pm$16.7 \\ 18.0 [4.0--30.0]} &
N/A &
Varying &
173.6 &
152.7 &
\textcolor{myyellow}{DUA-restricted} \\
 
\hline
\textbf{Total Overall} &
\textbf{3135} &
{--} &
\makecell{\textbf{29.8$\pm$17.7} \\ \textbf{28.0 [14.0--42.0]}} &
\makecell[l]{\textbf{Female: 50.3\%} \\ \textbf{Male: 41.1\%} \\ \textbf{Missing: 8.5\%} \\ \textbf{Non-binary: 0.1\%}} &
\makecell{\textbf{146.8$\pm$50.7} \\ \textbf{150.0 [120.0--180.0]}} &
\makecell{\textbf{16.3$\pm$13.4} \\ \textbf{13.0 [5.0--24.1]}} &
\makecell[l]{\textbf{G6: 64.6\%, Miss.: 18.1\%} \\ \textbf{G4: 7.8\%, G5: 4.1\%} \\ \textbf{Guard.3: 3.6\%, Libre 2: 0.9\%} \\ \textbf{Enlite: 0.4\%, others $<$0.4\%}} &
{--} &
\textbf{2030.1} &
\textbf{1228.2} &
{--} \\
\hline
\end{tabular}
\end{table}

\end{landscape}

\restoregeometry

\subsection{MetaboNet Data Format}

The dataset follows the data harmonization conventions as illustrated in Figure \ref{fig:fig_1_v2}, with detailed descriptions elaborated in previous work \cite{WolffRoystonFougner2025HarmonizingDiabetesDatasets}. This includes resampling the datasets into a tabular format, with a homogeneous 5-minute time grid, and standardizing units and feature names. Each row corresponds to a unique subject and timestamp pair, and all records are consolidated into a single file. CGM measurements, insulin delivery records, date information, source file identifiers, and subject identifiers are available for all participants. The core features include CGM readings, insulin data, and meal information. Insulin data are provided as basal, bolus, and total insulin, where total insulin represents the sum of basal and bolus. Variability exists across studies: some report only a merged insulin column, while participants on multiple daily injection (MDI) therapy always have basal and bolus values recorded separately. Meal information includes carbohydrate entries and user-reported meal labels describing the type or content of each meal. Additional features, such as exercise and physiological signals, are incorporated when available, resulting in differences in feature coverage across sources. A complete description of all features in the dataset is available on the MetaboNet website \cite{MetaboNetNDDataDictionary}.

All units are standardized, with CGM measurements reported in mg/dL. Insulin administration is separated into bolus and basal insulin where possible, reported in absolute insulin units. If source datasets had stored insulin delivery as rates, such as U/hr, total insulin delivered was calculated from the rate and duration, expressed as the absolute amount delivered within each 5-minute time interval. Insulin types, when available, are included in the features “insulin\_type\_bolus” and “insulin\_type\_basal”. For insulin pump users, these two features have identical values. A column “insulin\_delivery\_modality” indicates delivery type with values Multiple Daily Injections (MDI), Sensor Augmented Pump (SAP), or AID.

Ethnicities are harmonized using category names based on the U.S. 2000 classifications for race \cite{grieco2001race}: White, Black/African American, Asian, Native Hawaiian or Other Pacific Islander, and American Indian or Alaska Native. Multi-racial entries are comma-separated, while race and ethnicity are merged, with "Hispanic/Latino" added where applicable.  For datasets in other formats, we standardized the data by mapping the original values to the corresponding categories in this classification.

As the dataset represents real-world data, missing values are inherent and preserved, leaving downstream users to select an appropriate imputation strategy. Signals such as meals, insulin boluses, and physical activity are naturally sparse. Users should be aware that zero values and missing entries for carbohydrate intake and insulin delivery are not unambiguously distinguishable, as zeros may represent missing data and missing entries may correspond to no delivery. Moreover, features obtained through manual reporting, such as carbohydrate intake and exercise, may be prone to errors as they may be affected by inaccuracies inherent in human data entry.

Features representing cumulative quantities, such as insulin delivery and carbohydrates, are summed over the five-minute interval, whereas features representing rates, such as heart rate, are averaged to match the dataset frequency. Demographic information, including gender and ethnicity, is repeated for each row corresponding to the same subject.

\subsection{MetaboNet Quality Assurance}
MetaboNet includes data from various sources, where most were thoroughly cleaned prior to public release and incorporation into MetaboNet. After harmonizing the existing datasets, all datasets included in MetaboNet were validated using quality assurance tests, including range checks, removal of duplicate records, and verification that device times matched enrollment periods. If issues were identified in the raw data, the corresponding entries were discarded; if the issue arose during data processing, it was corrected. When the validity remained uncertain, values were evaluated in the context of demographics and other features to assess plausibility. For JAEB studies, any unexpected device data inconsistent with enrollment or visit dates, or any issues identified during range checks (e.g., device data, laboratory measurements, case report forms), were investigated by JAEB.

Duplicate records may arise when patients participate in both a randomized controlled trial (RCT) and one or more observational studies during overlapping time periods.  Especially when intermingling datasets like the Loop Observational study and OpenAPS Commons, where the decision to donate data may have come long after data was initially collected, duplicate records are a concern.  We built on the prior work by Cooper et. al. on CGM deduplication \cite{CooperReinholdShahid2025GlucoseVariabilityTwoDatasets}, which focuses on matching statistics computed on CGM metrics on a per-day basis. Ultimately, we found 788 duplicated days of data across 5 patients. Only the duplicates were removed, so in these 5 instances, a patient's data might be spread across multiple user ids.

\section{Results}
\subsection{Broad Coverage}
\begin{figure}
    \centering
    \begin{tikzpicture}
        \begin{axis}[
            width=15cm, height=9cm,
            title={\textbf{Dataset Overview}},
            xlabel={Number of Subjects},
            ylabel={Patient Years of CGM and Insulin Data},
            xmin=0, xmax=3750,
            ymin=0, ymax=1300,
            grid=both,
            major grid style={line width=0.2pt, draw=gray!40},
            minor grid style={line width=0.1pt, draw=gray!15},
            legend style={
                at={(0.985,0.985)},
                anchor=north east,
                draw=black,
                fill=white,
                font=\small,
                row sep=2pt,
            },
            legend cell align={left},
            every axis plot/.append style={area legend},
            tick align=outside,
        ]
         
         
        \addplot[
            pattern=north east lines,
            pattern color=blue!70!black,
            draw=blue!70!black,
            line width=0.8pt,
        ] coordinates {(0,0) (3135,0) (3135,1228.2) (0,1228.2)} \closedcycle;
        \addlegendentry{Total Dataset Combined}
         
        \addplot[fill=white, draw=none, forget plot]
            coordinates {(0,0) (1895,0) (1895,867.4) (0,867.4)} \closedcycle;
        \addplot[
            pattern=crosshatch dots,
            pattern color=green!45!black,
            draw=green!45!black,
            line width=0.8pt,
        ] coordinates {(0,0) (1895,0) (1895,867.4) (0,867.4)} \closedcycle;
        \addlegendentry{Public Dataset}
         
        \addplot[fill=white, draw=none, forget plot]
            coordinates {(0,0) (1240,0) (1240,360.8) (0,360.8)} \closedcycle;
        \addplot[
            pattern=horizontal lines,
            pattern color=orange!90!black,
            draw=orange!90!black,
            line width=0.8pt,
        ] coordinates {(0,0) (1240,0) (1240,360.8) (0,360.8)} \closedcycle;
        \addlegendentry{DUA-Governed Dataset}
         
        \addplot[fill=white, draw=none, forget plot]
            coordinates {(0,0) (738,0) (738,36.9) (0,36.9)} \closedcycle;
        \addplot[
            pattern=crosshatch,
            pattern color=purple!80!black,
            draw=purple!80!black,
            line width=0.8pt,
        ] coordinates {(0,0) (738,0) (738,36.9) (0,36.9)} \closedcycle;
        \addlegendentry{T1DEXI and T1DEXIP}
         
        \end{axis}
        \end{tikzpicture}
    \caption{Overview of the scale of MetaboNet 2026, compared with the T1DEXI dataset after preprocessing according to the study’s inclusion criteria. The green portion represents publicly available data, orange indicates datasets governed by data use agreements (DUAs), and blue corresponds to the combined public and DUA-protected datasets. Patient-years of CGM and insulin data are defined as periods during which continuous glucose monitoring and insulin dosing are recorded.}
    \label{fig:1}
\end{figure}
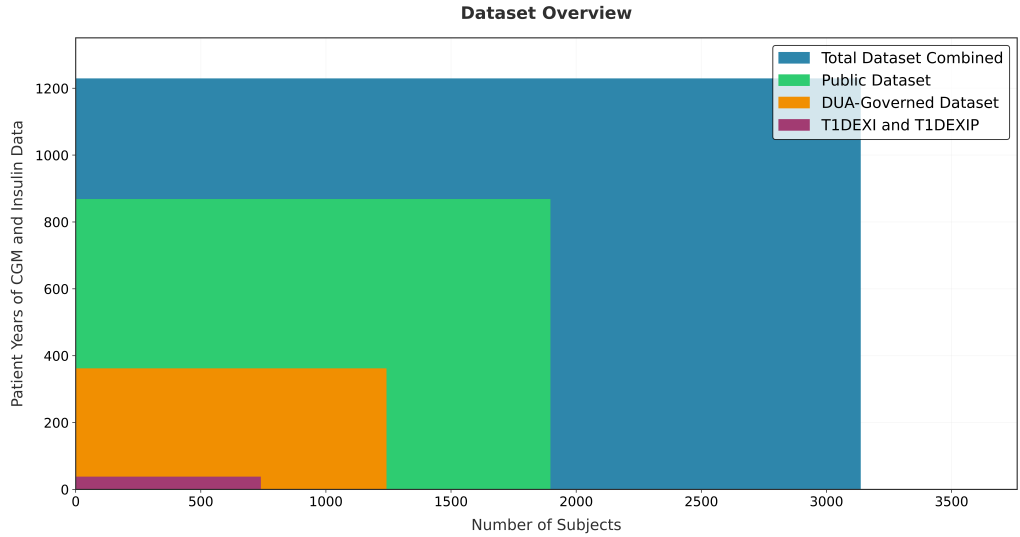

The resulting combined dataset comprises 3135 subjects and 1228 patient-years of data, calculated based on periods where both CGM values and non-zero insulin values are available. As illustrated in Figure \ref{fig:1}, the combined data are substantially larger than any individual dataset, with T1DEXI and T1DEXIP together as the reference point. This reference point was chosen because it is a widely used dataset \cite{RiddellLiGal2023T1DEXI,ChoAielloOzaslan2024PhysicalActivityDetectionFramework}, and one of the largest within MetaboNet in terms of subject cohort and patient-years. The size of MetaboNet is important for robust estimation of intra- and inter-individual variability of insulin-glucose dynamics. The publicly available data represents 71{\%} of the total patient-years of overlapping CGM and insulin data.

\begin{figure}
    \centering
    \includegraphics[width=0.97\linewidth]{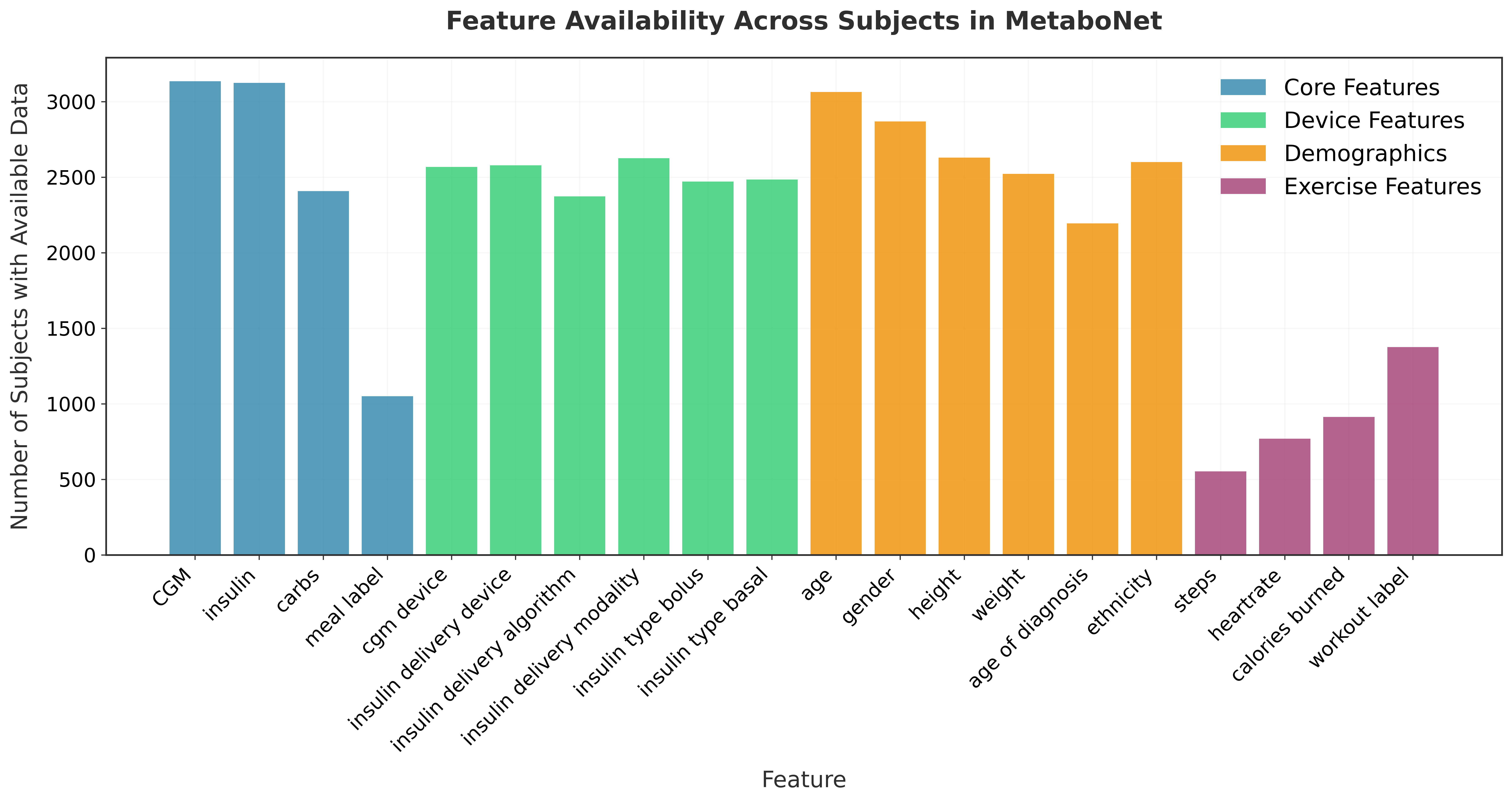}
    \caption{Subject-level feature availability across the MetaboNet dataset. Each bar represents the number of subjects for which at least one non-missing value is available for the corresponding feature. This figure includes a subset of features, while the full list of available features is provided on the MetaboNet website \cite{MetaboNetNDDataDictionary}.}
    \label{fig:2}
\end{figure}

MetaboNet also covers a broad range of features. Figure \ref{fig:2} presents the subject-level feature coverage for a subset of MetaboNet features, showing the number of subjects with at least one non-missing value for the corresponding feature. A detailed overview of feature coverage across the source datasets is presented in Figures \ref{fig:S1_v2} and \ref{fig:S2_v2}. Several core, demographic, and device-related features are available for at least 2,000 participants, i.e. 60{\%} of the cohort. The broad coverage of demographic and device-related features enables stratified analyses across population subgroups. Among the physical activity-related measures, the workout label is the feature that spans the most subjects, with available data for approximately 1,400 participants. Although the physical-activity-related features are more limited, they are still present in a significant proportion of the dataset and offer opportunities to explore new hypotheses. The availability of physical activity indicators is essential to address the unmet need for personalized strategies to mitigate physical activity-related hypoglycemia in T1D \cite{DhaliwalTangAiello2025HypoglycemiaRiskPhysicalActivity}.

\begin{figure}
    \centering
    \includegraphics[width=0.97\linewidth]{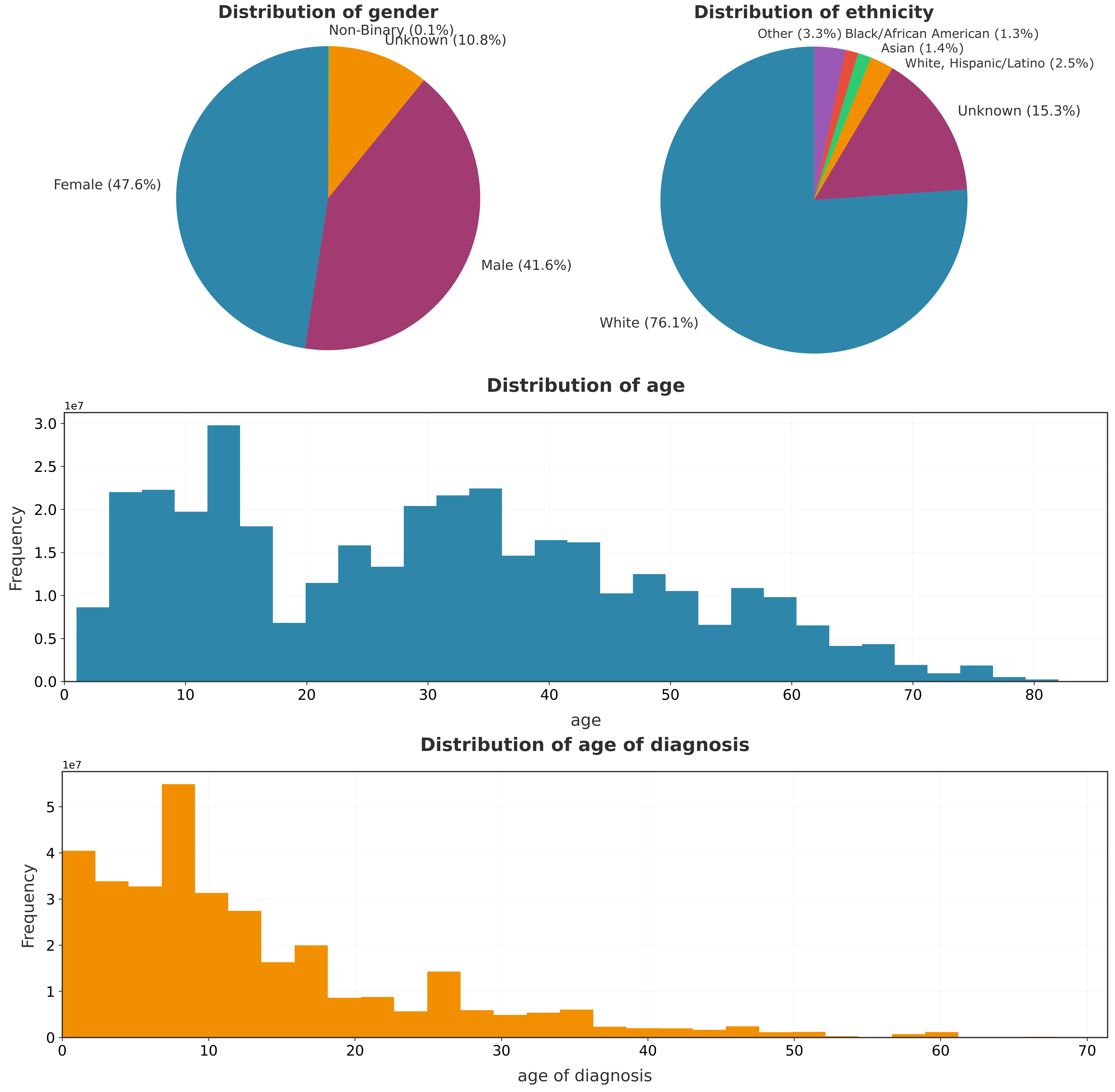}
    \caption{Demographic distribution of the dataset. The top panels show the proportion of individuals by gender (left) and ethnicity (right), with the majority identifying as female and white, respectively, and a notable fraction in the “unknown” category for both attributes. The bottom panel displays the age and age of diagnosis distributions.}
    \label{fig:3}
\end{figure}

\begin{figure}
    \centering
    \includegraphics[width=0.97\linewidth]{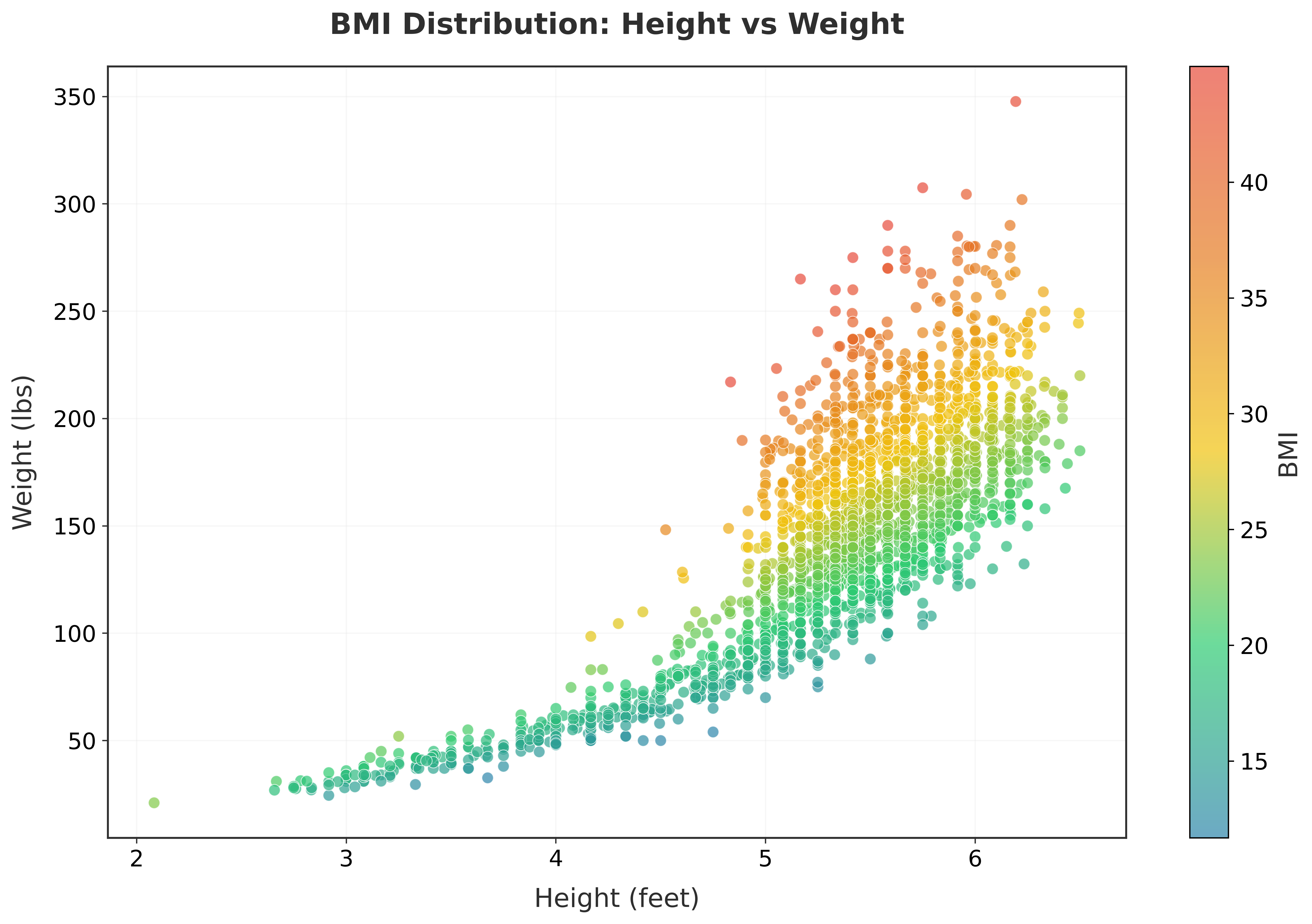}
    \caption{Scatter plot showing the relationship between height (x-axis) and weight (y-axis). Each point represents one subject in the full dataset, with point colour indicating the individual’s BMI category. Because pediatric BMI classification requires age- and sex-specific criteria defined by the World Health Organization BMI-for-age growth references, we report raw BMI values and avoid fixed weight-status categories.}
    \label{fig:4}
\end{figure}

Figure \ref{fig:3} indicates that the dataset encompasses a broad demographic spectrum, with a balanced gender distribution and a wide range of ages represented. The age distribution is right-skewed; this pattern is expected given typical participation rates in population studies, where older adults are less likely to enroll. The dataset also includes multiple ethnicities. While individuals of White ethnicity are overrepresented, this reflects the demographics of the contributing studies and source countries. Importantly, the presence of multiple ethnic groups still enables preliminary subgroup analyses and underscores the need for future dataset expansion to improve representation. Figure \ref{fig:4} shows that participants span a wide range of body-mass index (BMI) values, further demonstrating heterogeneity in key demographic and physiological characteristics. This heterogeneity is important for evaluating predictive models, as glycemic dynamics and insulin requirements vary with age, sex, and body composition.

Overall, Figures \ref{fig:1}-\ref{fig:4} show that this dataset captures substantial variability across dimensions. Figures \ref{fig:supp-2}-\ref{fig:supp-7} further characterize MetaboNet by presenting the per-study distributions of Total Daily Dose (TDD), CGM, Body Mass Index (BMI), and T1D duration, as well as the overall distributions. These figures highlight variability in these key features across studies, underscoring the limitations of relying on single cohorts for generalizable conclusions. By consolidating these datasets, MetaboNet captures broader clinical and physiological variability, reduces study-specific bias, and enables more robust population-level inference.

\subsection{Applications}
MetaboNet’s diverse and longitudinal dataset enables population-level analyses of glycemic patterns across demographic and lifestyle subgroups. Insights gained from these analyses can inform the development, validation, and benchmarking of algorithms that are more generalizable across populations. In this section, we present two examples of such applications.

\subsubsection{Population Level Analyses}
Recent research highlights the need for investigating CGM-derived metrics \cite{PionaAielloMancioppi2025PrediabetesYouthObesity,DunnAjjanBergenstal2024BeyondTIRtoTITR,BeckRaghinaruCalhoun2024TIRvsTITR}, however, the sample size of the analyzed data is a limitation to the strength of the results. The MetaboNet dataset can further assess relationships among CGM-derived metrics using a large database. Notably, the MetaboNet dataset captures a wide spectrum of glycemic control, from participants with consistently tight glucose regulation to those with more variable levels, enabling analyses across diverse profiles. As an example, we examined the relationship between the percentage of TIR 70--180 mg/dL (3.9--10 mmol/L) and the percentage of Time in Tight Range (TITR) 70--140 mg/dL (3.9--7.8 mmol/L). A scatter plot of the percentage of TIR versus TITR for each participant with at least 14 days of continuous glucose monitoring data is provided in Figure \ref{fig:supp-1}. This analysis illustrates only one out of several potential large-scale population-level analyses enabled by MetaboNet, as well as the diversity of glycemic profiles among the individuals in the dataset. 

\subsubsection{Data-Driven Algorithm Development}

MetaboNet provides a valuable resource for training and validating data-driven algorithms in diabetes management. One prominent application is glucose prediction. A recent systematic review of deep learning algorithms for glucose forecasting in type 1 diabetes by Calzavara et al. found that cross-study performance comparisons across different datasets and preprocessing techniques remain a key issue \cite{calzavara2025systematic}. They also found that nearly half of the studies used the OhioT1DM dataset, which accounts for approximately 0.1\% of the public version of MetaboNet in terms of total data days. This supports the need for standardized, large-scale datasets to enable more generalizable benchmarking of advanced algorithms.

We illustrate this using a benchmark approach in the Supplementary Material comparing naïve, linear, and non-linear models over a 30-minute prediction horizon. Model descriptions, data processing and partitioning, and performance metrics are covered in the Supplementary Material. The results illustrate that increasing the amount of training data improves prediction accuracy on the test set (Figure \ref{fig:S10}). These findings suggest the potential of MetaboNet to support machine learning research and development in diabetes management. However, data-driven black box modeling does not necessarily guarantee causal inference, especially when trained on large datasets. Additional model validation is essential to ensure that the model's validity does not result in spurious correlations.

\section{How to Access MetaboNet}
The public part of the dataset can be accessed via \url{https://metabo-net.org}. The user must log in to access the public data directly. Data can be downloaded as either a single parquet file or as separate files for each individual dataset.

The DUA-governed datasets cannot be accessed through the MetaboNet website. Each dataset must be applied for through the dataset's respective application process. When the raw data is accessed, a published open-source code is provided to process it into the MetaboNet format (available at \url{https://github.com/replicahealth/metabonet_processor}). Follow the instructions in the code repository to process the raw DUA-governed datasets into the MetaboNet format.

In the consolidated dataset file download on the MetaboNet website, a dropdown menu lists the available dataset versions, with the latest version as the default. When using the dataset in an academic publication, we recommend indicating which version is used in the manuscript to enhance reproducibility. All of the figures and tables in this manuscript have been generated using MetaboNet version 1.0.

\section{Discussion}
The MetaboNet dataset represents a significant step toward enabling large-scale, data-driven research in diabetes management, including but not limited to algorithm development and population-level analyses, with a large feature set. In this current version of MetaboNet (Metabonet 2026), 21 existing datasets were harmonized into a unified resource. Each independent dataset was originally collected for specific purposes and is therefore potentially subject to individual study biases. By consolidating these datasets, MetaboNet provides a more generalizable resource by capturing diverse subject profiles. Additionally, MetaboNet can help researchers save time and promote reproducible research by minimizing variability in data handling across independent studies.

MetaboNet supports conventional analyses using widely adopted CGM measures such as TIR, Time Below Range, mean glucose, and glycemic variability indices \cite{rand2026cgmreview}. Beyond these traditional metrics, the dataset enables population-level investigations of trends, variability, and subgroup differences across diverse patient profiles. Established areas of research such as glucose prediction, hypoglycemia forecasting \cite{bergford2023exercise,leutheuser2024nocturnal,piersanti2023ml,cinar2025benchmarking}, and meal detection algorithms \cite{TurksoySamadiFeng2016MealDetectionModule} have largely relied on smaller datasets, and could benefit from validation and benchmarking on a harmonized resource like MetaboNet.

The dataset also provides opportunities to apply and evaluate state-of-the-art techniques that require large and heterogeneous datasets. Traditional CGM analytics can be expanded to high-dimensional analytics such as glucodensity-based representations \cite{matabuena2025glucodensity} or functional data analysis including multilevel functional models \cite{matabuena2026functional}, by capturing complex temporal patterns, distributions, and individual variability in glucose dynamics. MetaboNet can further support development and testing of novel diabetes management strategies using off-policy evaluation (OPE), which allows retrospective assessment of multiple control algorithms from observational data \cite{FuNorouziNachum2021BenchmarksDeepOPE}. Data-driven approaches such as offline reinforcement learning (RL) \cite{EmersonGuyMcConville2023OfflineRLSaferBG,ZhuLiGeorgiou2023OfflineDRLBasalInsulin} can be explored within this framework. Across this spectrum, from widely used CGM metrics to sophisticated modeling and control strategies, robust evaluation and generalization benefit from a large, diverse dataset encompassing multiple devices, demographic groups, and state–action spaces \cite{LevineKumarTucker2020OfflineRLTutorial}.

A limitation of our study is that datasets were not identified through a systematic review, and therefore, some relevant datasets may not have been included. Additionally, temporal variability presents challenges, as time zone changes and irregular measurement schedules may introduce artifacts. While resampling and filtering help mitigate these effects, some inconsistencies may persist due to the dataset’s scale and heterogeneous preprocessing across sources. Dataset diversity is a key strength, as it enables analyses across heterogeneous populations. However, integrating heterogeneous data sources may introduce structural and measurement-related biases. The contributing datasets differ in recruitment strategies, inclusion criteria, devices, sampling frequencies, study protocols, and follow-up durations, potentially generating systematic differences in outcomes unrelated to the phenomenon of interest. For example, heart rate and glucose measurements may be collected using different hardware platforms that have already undergone device-specific preprocessing and therefore may not represent equivalent signals when integrated. Hence, it is essential for researchers utilizing MetaboNet to take into account the characteristics of the contributing studies when designing analyses and interpreting results, in order to address potential systematic differences and ensure valid conclusions.

Future efforts will focus on expanding and enhancing this harmonized dataset to improve its translational value in Type 1 Diabetes research. Potential directions include augmenting the dataset with longer-term clinical outcomes, such as HbA1c trajectories and diabetes-related complications. While some source datasets already include these features, others could explore deriving them from existing data, such as prediction of HbA1c from CGM measurements \cite{qaraqe2024fewshot}. Further expansions, such as integrating individuals with Type 2 Diabetes, could increase the dataset’s value for broader research applications. Finally, future efforts will focus on establishing standardized DUA and centralized workflows to enhance accessibility and reproducibility. This approach aims to reduce friction for researchers while maintaining governance requirements and supporting rigorous, reproducible analyses.

\section{Conclusion}
In conclusion, MetaboNet provides a comprehensive and diverse resource for data-driven research in diabetes management. The dataset consistently includes CGM and insulin information, and, when available, also incorporates data on food intake, physical activity, device characteristics, and patient demographics. By integrating data from multiple studies, MetaboNet offers a large cohort with broad participant representation, supporting more generalizable analyses and enabling the investigation of a wide range of research questions, particularly those related to data-driven algorithm development.

A substantial portion of the harmonized dataset is publicly accessible and immediately downloadable, whereas access to the remaining data requires an application to the respective data owners. To facilitate the integration of these DUA-governed datasets, we provide an open-source processing pipeline that standardizes heterogeneous sources, thereby enhancing reproducibility through consistent data-handling procedures. We encourage the research community to contact us regarding additional datasets that may be integrated into future releases of MetaboNet, thereby expanding its scope and enhancing its potential for scientific discovery and personalized diabetes care.

\section*{Acknowledgements}

We would like to thank all individuals, institutions, and organizations that made the datasets used in this study publicly available. Their contributions to data collection and sharing were essential for the completion of this work. We thank Courtney O'Donnell for her development work contributing to https://metabo-net.org/.

This publication is based on research using data from the Jaeb Center for Health Research, retrieved from https://public.jaeb.org/datasets/diabetes. The analysis's content and conclusions presented herein are solely the responsibility of the authors and have not been reviewed or approved by the source dataset study group or study sponsor. 

This publication is based on research using data from the Type 1 Diabetes EXercise Initiative (T1DEXI) Study that has been made available through Vivli, Inc. Vivli has not contributed to or approved, and is not in any way responsible for, the content of this publication.

\clearpage
\appendix
\section*{Supplementary Materials}
\addcontentsline{toc}{section}{Supplementary Materials} 

\renewcommand{\thesection}{S\arabic{section}}
\setcounter{section}{0}

\subsection*{Abbreviations} 

Continuous Glucose Monitoring (CGM), Body-Mass Index (BMI), Clarke Error Grid (CEG), Zero-Order Hold (ZOH), Linear Extrapolation (LE), Autoregressive (AR),  Physiology-Based Design (PBM), Insulin Sensitivity Factor (ISF), Carbohydrate Ratio (CR), Total Daily Dose (TDD), Neural Network (NN), Support Vector Regression (SVR), Radial-Basis Function (RBF), Multilayer Perceptron (MLP)

\section{The Broad Coverage of MetaboNet 2026}

The feature space varies across the source study datasets (Figure \ref{fig:S1_v2}), and the inclusion of specific variables in MetaboNet depends on their availability within each study. As these are real-world datasets, they exhibit varying degrees of missingness both across and within features. CGM coverage differs substantially between sources (Figure \ref{fig:S2_v2}): some datasets show consistently high coverage with low inter-participant variability, whereas others exhibit marked heterogeneity. These differences partly reflect variations in sampling frequency and preprocessing. For example, the ShanghaiT1DM dataset uses a 15-minute sampling interval, corresponding to 33\% coverage when harmonized to the 5-minute grid in MetaboNet, while the HUPA-UCM dataset applied linear interpolation to impute missing values, resulting in nominal 100\% coverage.

\setcounter{figure}{0}
\renewcommand{\thefigure}{S\arabic{figure}}

\begin{figure}[!ht]
    \centering
    \includegraphics[width=1\linewidth]{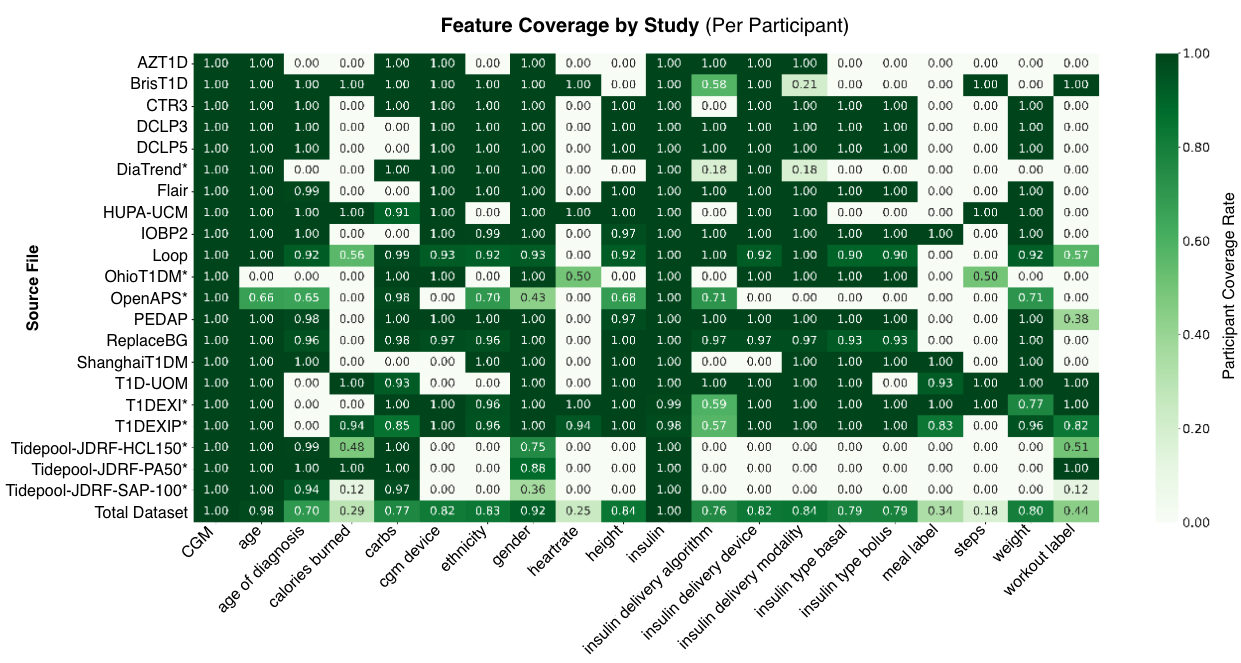}
    \caption{Per-participant feature coverage by study, for a subset of the features within MetaboNet. The heatmap shows the fraction of participants per study with at least one valid value within each feature. The asterisk (*) indicates DUA-governed datasets.}
    \label{fig:S1_v2}
\end{figure}

\begin{figure}[!ht]
    \centering
    \begin{minipage}[c]{0.7\linewidth}
        \centering
        \includegraphics[width=\linewidth]{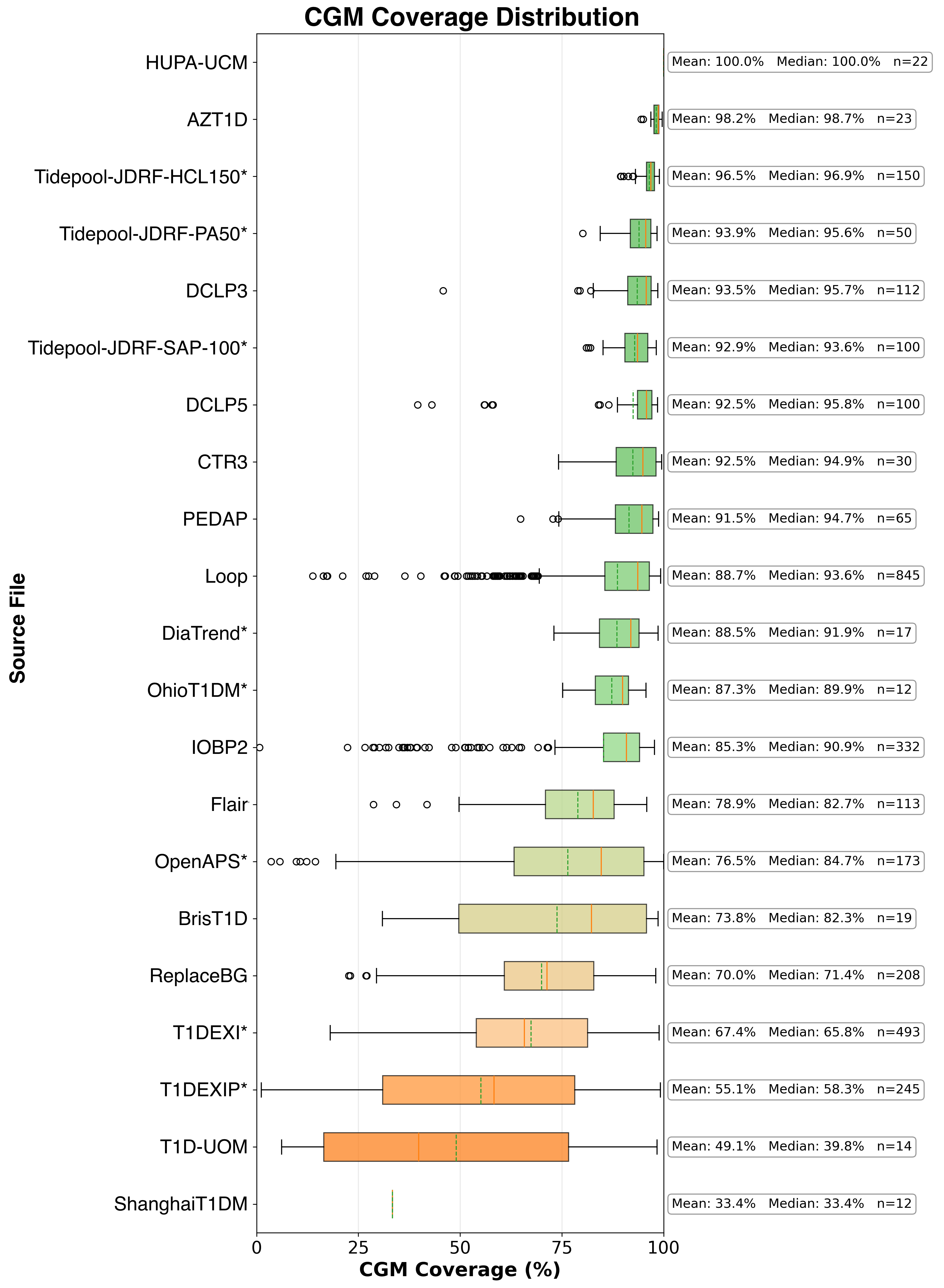}
    \end{minipage}\hfill
    \begin{minipage}[c]{0.28\linewidth}
        \caption{CGM data coverage variability by study. Horizontal box plots showing the distribution of participant-level CGM coverage percentages. Studies are ordered by the mean coverage, with colour coding from red (poor) to green (excellent coverage). The asterisk (*) indicates DUA-governed datasets.}
        \label{fig:S2_v2}
    \end{minipage}
\end{figure}

\begin{figure}
    \centering
    \includegraphics[width=1.0\linewidth]{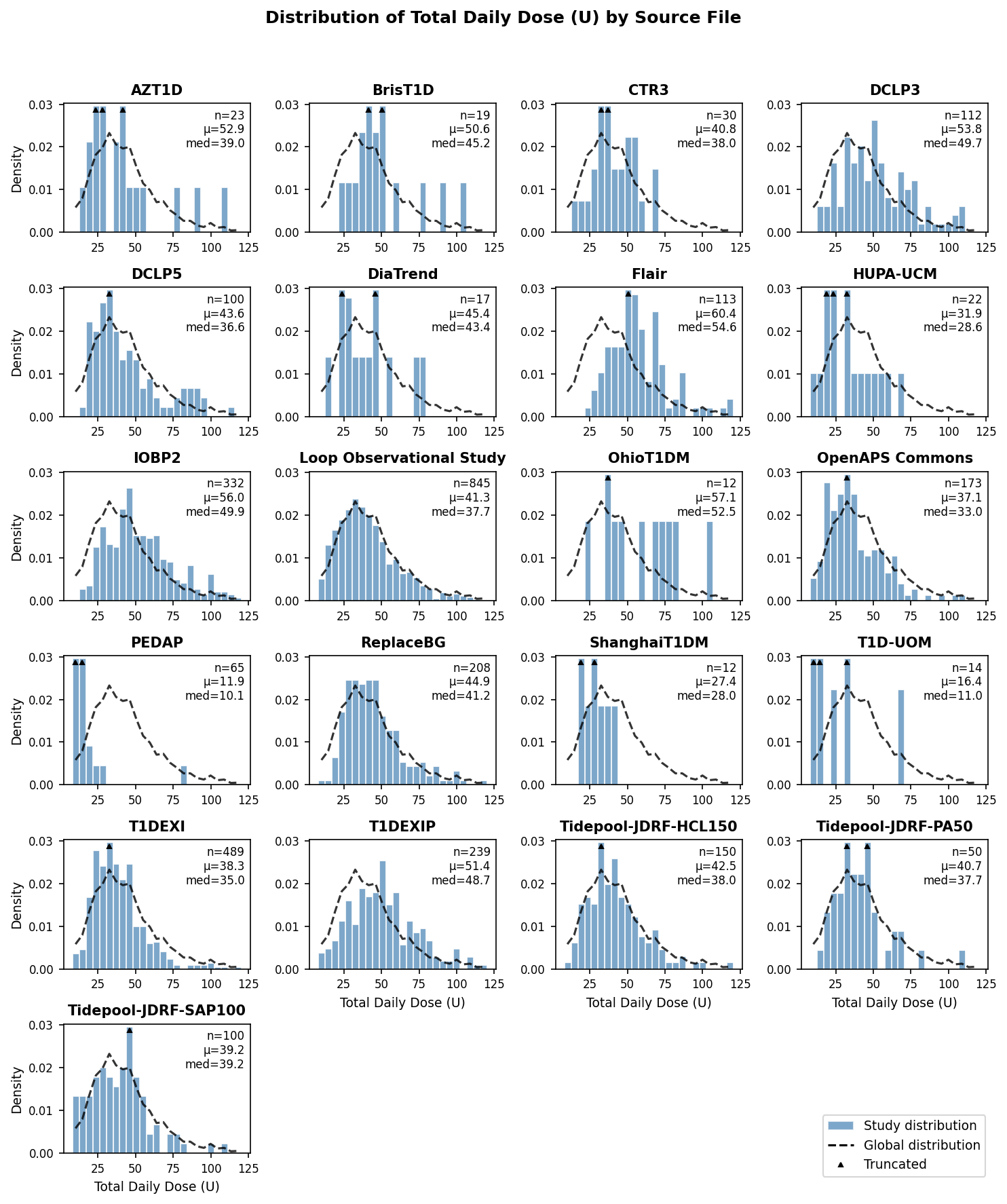}
    \caption{Distribution of total daily doses (TDD) per subject, shown separately for each study included in MetaboNet 2026. The dotted black line represents the distribution for the full dataset. For each distribution, the sample size, the mean ($\mu$), and the median values are reported.}
    \label{fig:supp-2}
\end{figure}

\begin{figure}
    \centering
    \includegraphics[width=1.0\linewidth]{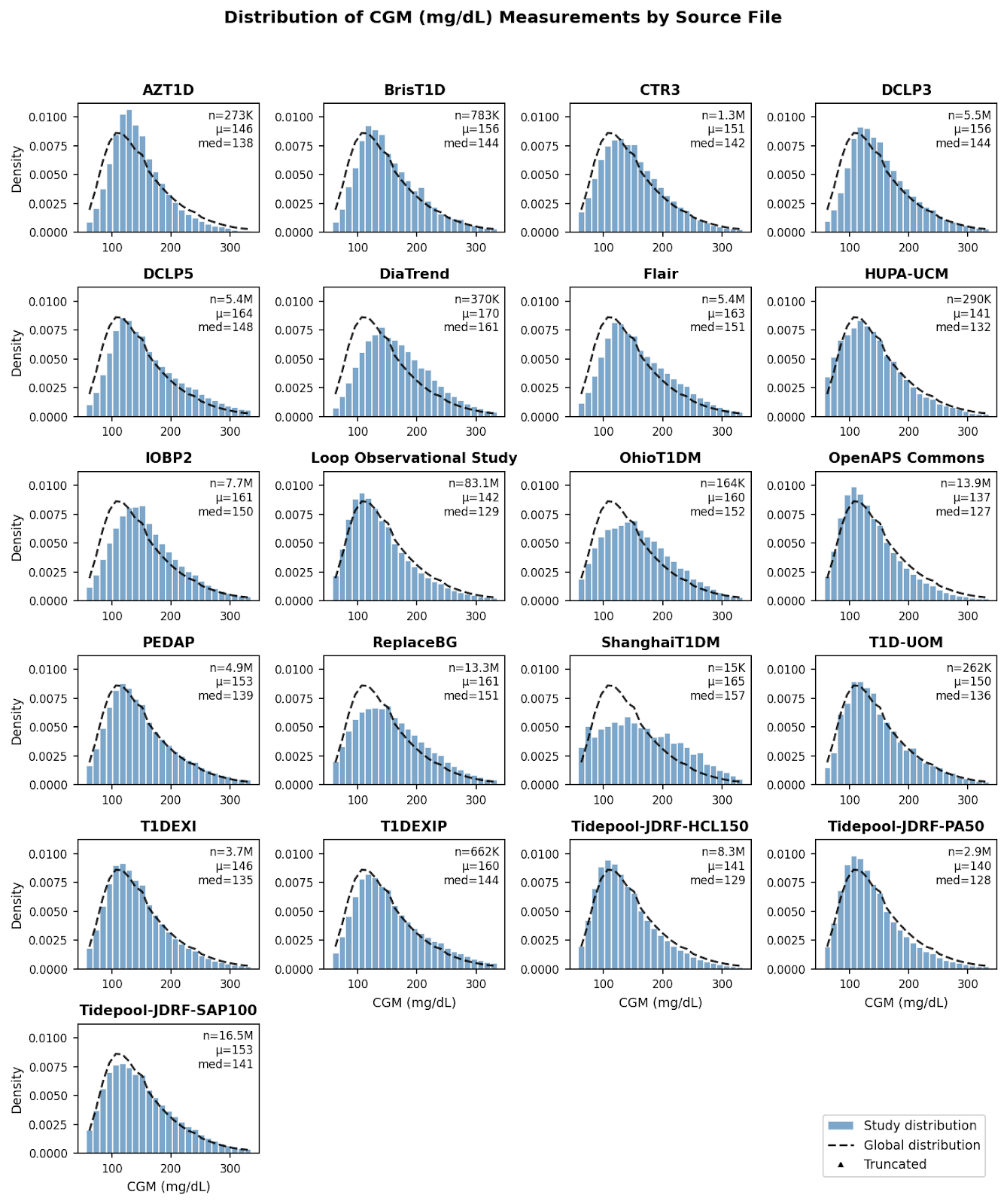}
    \caption{Distribution of Continuous Glucose Monitoring (CGM) measurements, shown separately for each study included in MetaboNet 2026. The dotted black line represents the distribution for the full dataset. For each distribution, the sample size, the mean ($\mu$), and the median values are reported.}
    \label{fig:supp-3}
\end{figure}

\begin{figure}
    \centering
    \includegraphics[width=1.0\linewidth]{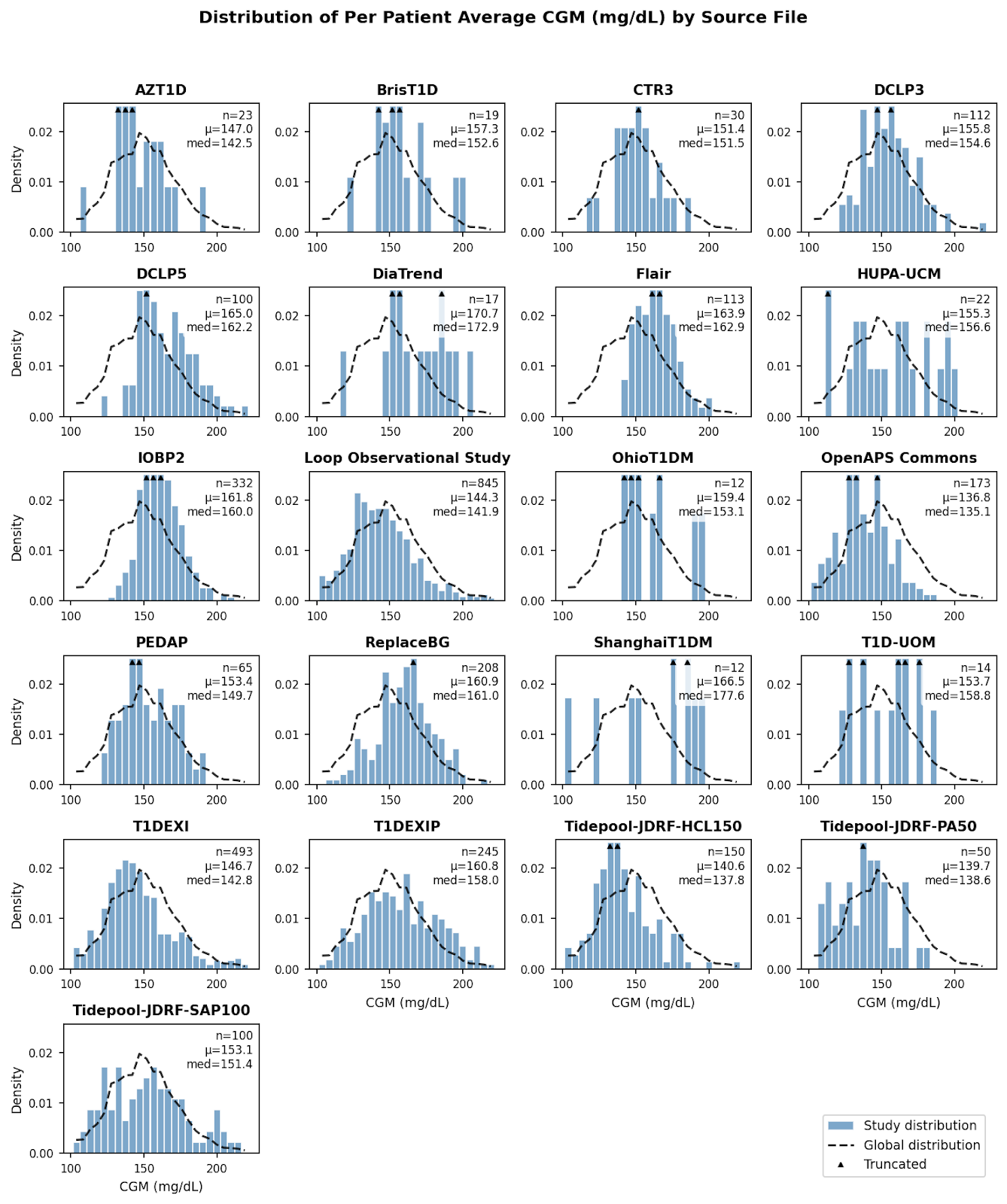}
    \caption{Distribution of per-patient average CGM, shown separately for each study included in MetaboNet 2026. The dotted black line represents the distribution for the full dataset. For each distribution, the sample size, the mean ($\mu$), and the median values are reported.}
    \label{fig:supp-4}
\end{figure}

\begin{figure}
    \centering
    \includegraphics[width=1.0\linewidth]{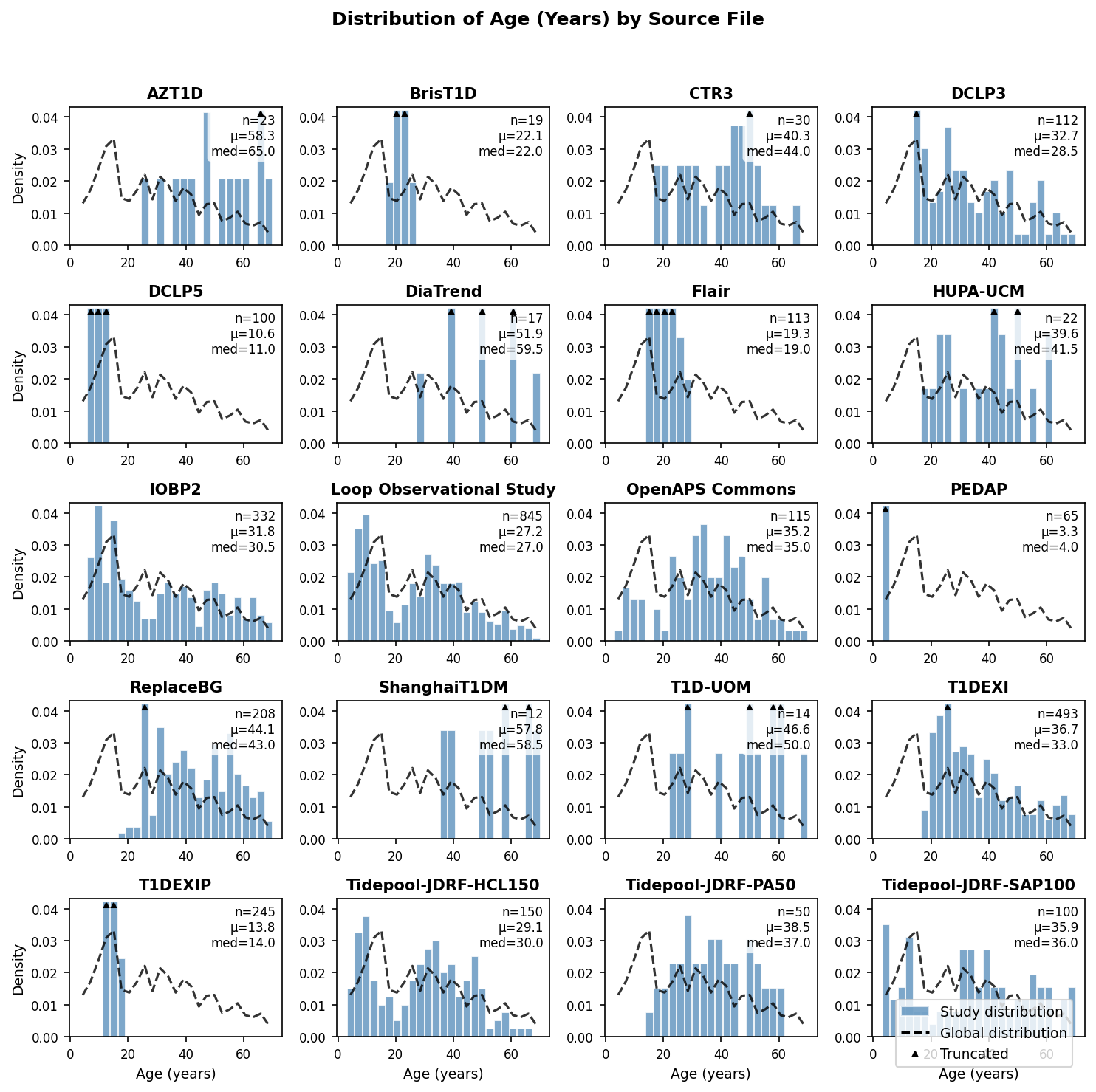}
    \caption{Distribution of patient age, shown separately for each study included in MetaboNet 2026. The dotted black line represents the distribution for the full dataset. For each distribution, the sample size, the mean ($\mu$), and the median values are reported. OhioT1DM did not include age data, and is therefore not visible here.}
    \label{fig:supp-5}
\end{figure}

\begin{figure}
    \centering
    \includegraphics[width=1.0\linewidth]{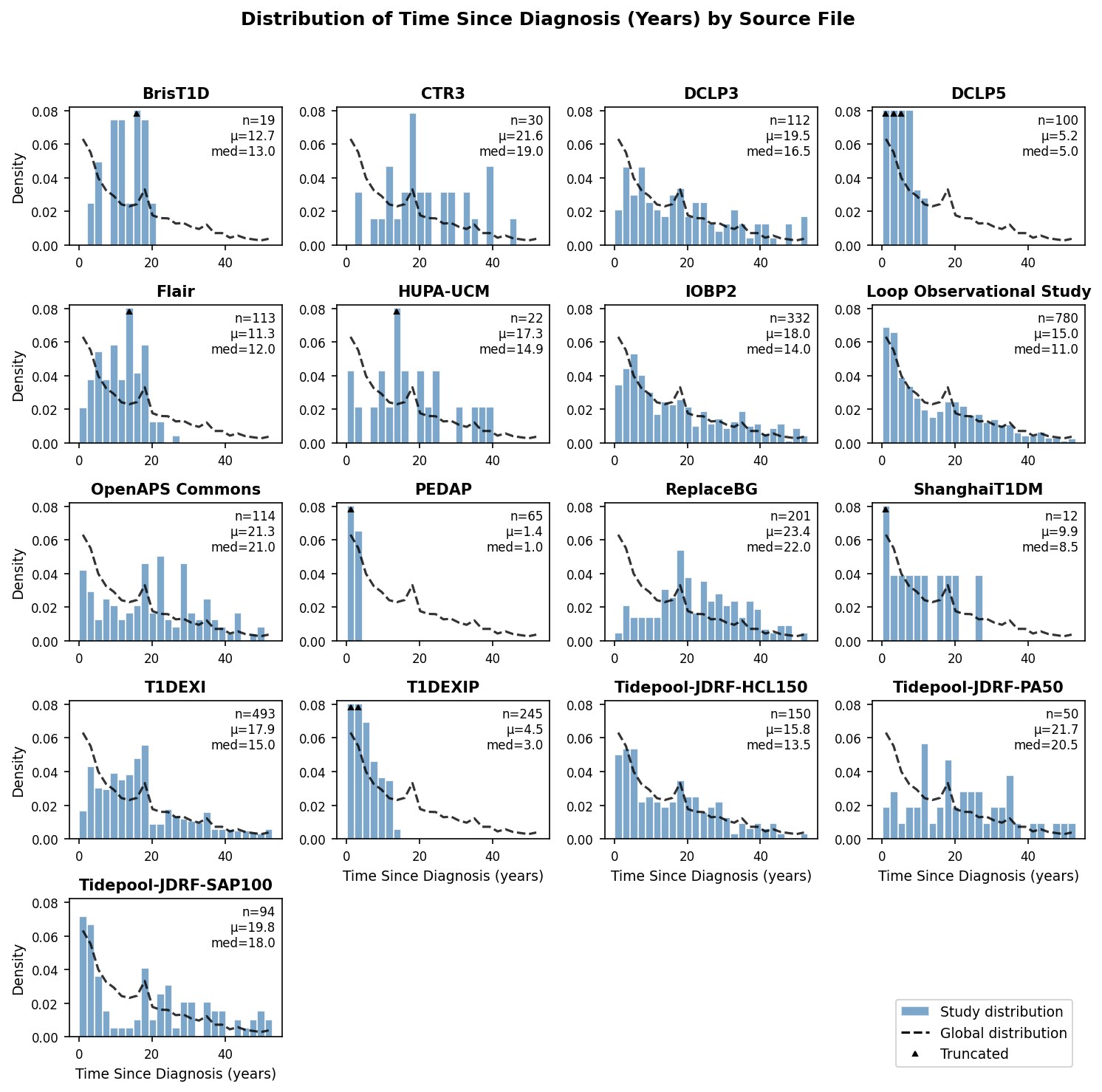}
    \caption{Distribution of T1D duration. This measure requires both age and age of diagnosis to be present in the source file, which excludes Diatrend, OhioT1DM, AZT1D, and T1D-UOM from this figure. For each distribution, the sample size, the mean ($\mu$), and the median values are reported.}
    \label{fig:supp-6}
\end{figure}

\begin{figure}
    \centering
    \includegraphics[width=1.0\linewidth]{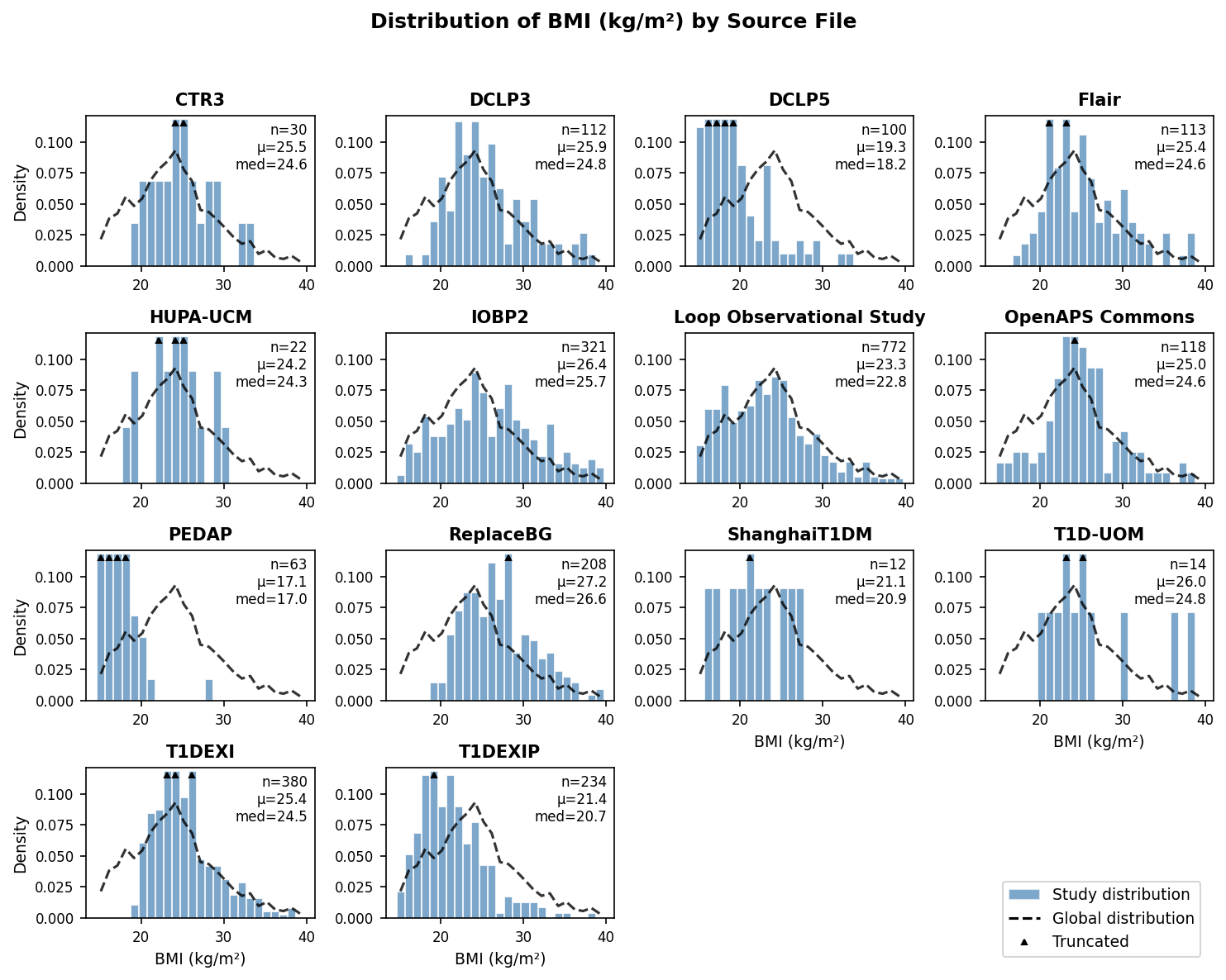}
    \caption{Distribution of Body-Mass Index (BMI).  This measure requires both ‘height’ and ‘weight’ to be present in the source file, which excludes Diatrend, OhioT1DM, the Tidepool data, T1DEXIP, AZT1D, and BrisT1D from this figure. For each distribution, the sample size, the mean ($\mu$), and the median values are reported.}
    \label{fig:supp-7}
\end{figure}

Figures \ref{fig:supp-2}–\ref{fig:supp-7} further characterize the dataset by presenting the per-study distributions of Total Daily Dose (TDD), Continuous Glucose Monitoring (CGM), Body Mass Index (BMI), and Type 1 Diabetes (T1D) Duration, as well as the overall distributions. For visual clarity, each axis is fixed, and outlying values are clipped in the top and bottom percentiles. Sample sizes may vary due to the availability of each variable in the study dataset. 

As shown in Figures \ref{fig:supp-2}–\ref{fig:supp-7}, each study exhibits its own characteristic distribution and potential biases, but collectively, they provide a comprehensive representation of glycemic profiles. TDD and T1D durations show broad global distribution with mean ± standard deviation values of 43.7 ± 23 IU and 15.5 ± 13 years, respectively. Across the combined cohorts, BMI ranges from 11.7 to 48.5 kg/m2, and participant ages range from 1 to 82 years. This diversity is critical for developing robust, generalizable data analyses and predictive algorithms, as it ensures that models are trained and evaluated on realistic, heterogeneous data. Note that this dataset includes participants under 19 years of age. Because pediatric BMI classification requires age- and sex-specific criteria defined by the World Health Organization BMI-for-age growth references, we report raw BMI values and do not apply fixed weight-status categories. Additionally, Figure \ref{fig:supp-4} shows the large variability in per-patient average CGM, which allows to leverage the dataset for the development of patient-tailored strategies.

\cleardoublepage
\newpage

\section{Population Level Analyses}
\begin{figure}[!ht]
    \centering
    \includegraphics[width=0.69\linewidth]{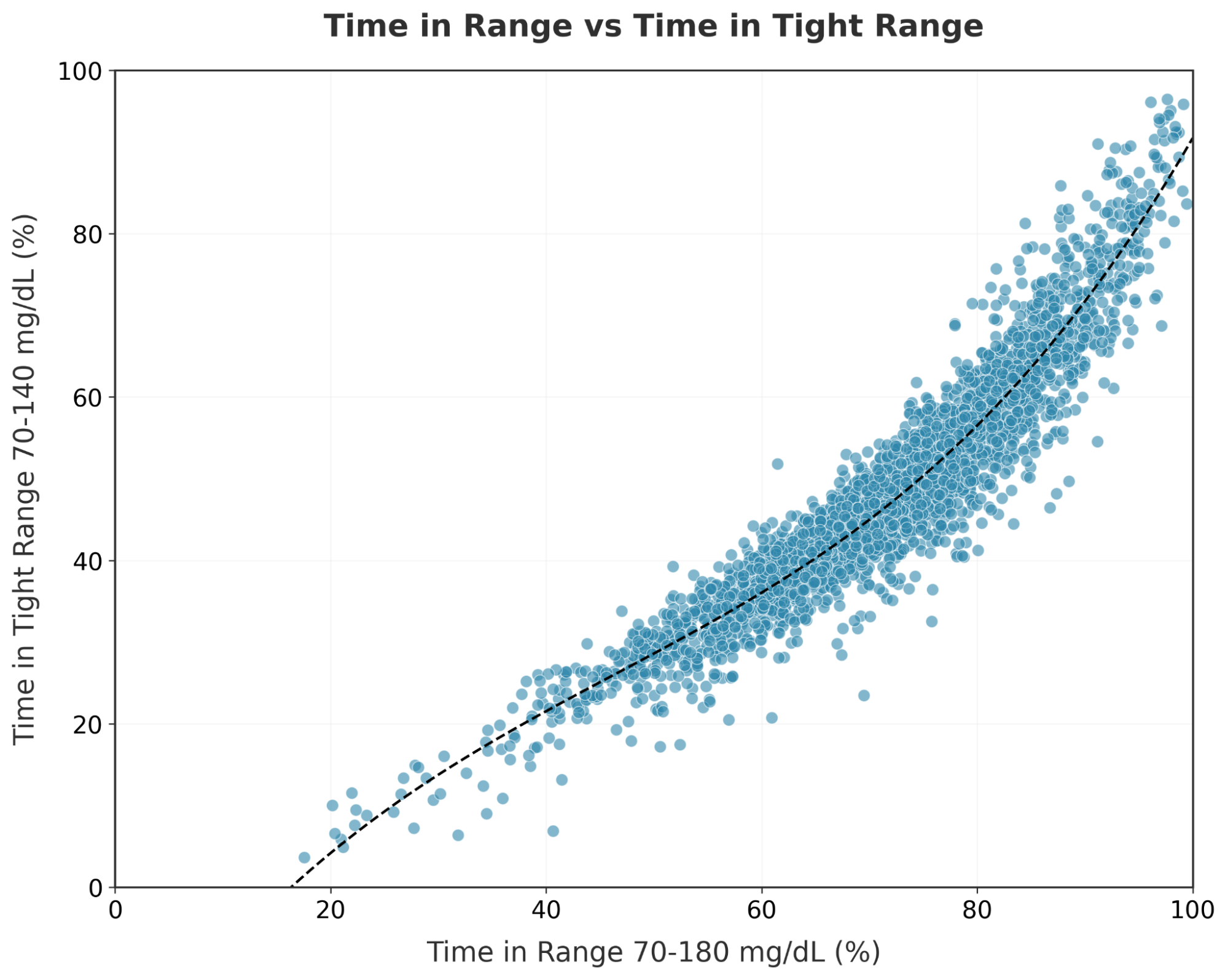}
    \caption{Relationship between the percentage of time in range (70–180 mg/dL) and the percentage of time in tight range (70–140 mg/dL) across participants included in MetaboNet 2026. Each point represents an individual subject, and only participants with at least 14 days of continuous glucose monitoring data are included. The dashed black line shows a third-degree polynomial fit using least squares, highlighting the overall relationship between the two metrics.}
    \label{fig:supp-1}
\end{figure}

MetaboNet 2026 encompasses a wide spectrum of glycemic profiles across its subjects. As illustrated in Figure \ref{fig:supp-1}, which presents Time in Range versus Time in Tight Range for each individual, the dataset includes subjects with glycemic control ranging from far below the clinical Time in Range target of 70{\%} to subjects exceeding that target by wide margins \cite{BattelinoDanneBergenstal2019TIRConsensus}. This highlights the diversity of glycemic behavior captured in the dataset, and shows the potential of MetaboNet for large-scale, population-level analyses.

\cleardoublepage
\newpage

\section{Blood Glucose Prediction Benchmarks}
\subsection{Models}

This section presents the blood glucose prediction models included in the benchmark. All models are evaluated using a 30-minute prediction horizon. The purpose of this benchmark is to provide a reference point for future studies using consolidated datasets. To capture the diversity of approaches commonly used in the literature, the models were selected to span key categories: simple naïve methods, a data-driven model with partial physiology-based design (PBD), support vector regression (SVR) \cite{HastieTibshiraniFriedman2009ESL}, a single-layer neural network (NN) \cite{HastieTibshiraniFriedman2009ESL}, and a Multilayer Perceptron (MLP) \cite{HastieTibshiraniFriedman2009ESL}. By covering baseline heuristics, physiologically inspired approaches, support vector regression, and neural networks, this selection represents several basic strategies employed in blood glucose prediction research.

\subsubsection{Naïve Models}
Recent consensus in the field recommends that blood glucose prediction algorithms be compared against simple baseline models, such as a zero-order hold (ZOH) predictor, a linear extrapolation (LE)  model, and a 3\textsuperscript{rd} order autoregressive (AR) model \cite{JacobsHerreroFacchinetti2023AIMLBestPractices}. 
The \textbf{ZOH} model is the simplest baseline, assuming that glucose levels will remain unchanged going forward:
\[
\hat{g}(k+1) = g(k)
\]
where $\hat{g}$ is the predicted glucose value, $g$ is the glucose measurement, and $k$ is the discrete time instant, and $g(k)$ is at $t=0$. The sampling time is equal to 5 minutes. 

The \textbf{Naïve LE} model improves slightly on this by extending the local linear trend of the most recent measurements. Including this model as a baseline aligns with the recommendations of Jacobs et al. in a consensus paper on best practices in artificial intelligence for diabetes \cite{JacobsHerreroFacchinetti2023AIMLBestPractices}. This model is also widely used as a baseline for fault detection applications, such as meal and physical activity detection. For each prediction time point, the glucose slope m at time k is estimated from the last three glucose values as the mean of the two consecutive differences. 
\[
m(k) = \frac{(g(k) - g(k-1)) + (g(k-1) - g(k-2))}{2}
\]

Let $g(k)$ be the glucose measured at time step $k$ (5-minute intervals). The 30-minute-ahead prediction $\hat{g}(k+6)$ is given by:
\[
\hat{g}(k+6)=g(k)+6m(k)
\]

Finally, the \textbf{Naïve AR} model uses a low-order temporal structure: a 3\textsuperscript{rd} order autoregressive formulation that predicts future glucose levels using the three most recent measurements, directly mapping those values to the predicted glucose level 30 minutes, i.e. 6-step ahead:
\[
\hat{g}(k+6) = \alpha_0 + \alpha_1 g(k) + \alpha_2 g(k-1) + \alpha_3 g(k-2)
\]
where $\alpha_0$, $\alpha_1$, $\alpha_2$, $\alpha_3$ are the model parameters, estimated using ordinary least squares. 

\subsubsection{Data-Driven Model with Partial Physiology-Based Design}
The \textbf{Data-Driven Model with Partial Physiology-Based Design (PBD)} incorporates both carbohydrate and insulin inputs and includes a retrospective correction module that adjusts predicted glucose trajectories based on the preceding 15 minutes. It is adapted from the Loop automated insulin delivery system and interfaces directly with its open-source implementation via a Python API (https://github.com/miriamkw/LoopAlgorithmToPython). To ensure physiological plausibility, predictions below 40 mg/dL and above 400 mg/dL are clipped at that threshold. Subject-specific therapy parameters—including basal insulin rate, insulin sensitivity factor (ISF), and carbohydrate ratio (CR)—are derived from the TDD, computed from the subject’s complete insulin record. Following common clinical heuristics, basal insulin is set to 45\% of the TDD \cite{ChawlaMakkar2019RSSDIInsulinConsensus}, ISF is calculated as 1800/TDD \cite{ChakrabartyZavitsanouDoyle2018EventTriggeredMPC, vanHeusdenDassauZisser2012ControlRelevantModels}, and CR as 500/TDD \cite{ChawlaMakkar2019RSSDIInsulinConsensus}. Notably, because the TDD is computed using insulin data from the entire record, this parameterization introduces information leakage from the future.

\subsubsection{Machine Learning Models}
The machine learning models share the same feature space, comprising CGM values, insulin, and carbohydrates. To capture temporal dynamics, time-lagged features are computed for the most recent two hours. The \textbf{Single Linear Layer NN} model is trained with backpropagation using PyTorch. One epoch, and a learning rate of 0.001 was used. The \textbf{SVR} model is implemented using stochastic gradient descent with L2 regularization and an $\epsilon$-insensitive loss function, enabling efficient iterative learning. This approach preserves the characteristics of an SVR with a linear kernel while allowing scalable updates suitable for large datasets. A benchmark for Xie et al. shows that an SVR with the linear kernel outperforms the radial-based function (RBF) kernel across various metrics \cite{XieWang2020BenchmarkingMLBGPrediction}. Lastly, the \textbf{MLP} model is a simple two-layer fully connected neural network with a ReLU-activated hidden layer of 64 units that maps a $d$-dimensional input to a single scalar output. This model was also trained for one epoch with a learning rate of 0.001.

\subsection{Data Processing}
The train-test split was designed so that each individual dataset within MetaboNet 2026 contributes to both training and testing. For each dataset, twenty percent of subjects were randomly assigned to the test set to guarantee the presence of previously unseen subjects relative to the training data. From the remaining subjects in the training pool, twenty percent were further split chronologically by time, with fifty percent allocated to training and fifty percent to testing, allowing some subjects to appear in both sets. A 24-hour buffer was applied between the splits to prevent information leakage.  Table \ref{tab:supp-1} summarizes the numbers of training and test samples after the processing. 

\setcounter{table}{0}
\renewcommand{\thetable}{S\arabic{table}}
\begin{table}[ht!]
\centering
\caption[Number of samples]{Number of samples after the train-test split and data processing, including imputation, time-lagged feature creation, and removal of rows with missing values. "Half observations" refers to subjects whose data are partially present in both the training and test sets, while "separate subjects" correspond to subjects included only in the test set.}
\begin{tabularx}{0.85\textwidth}{X S[table-format=8.0]}
\toprule
\textbf{Category} & \textbf{Number of Samples} \\
\midrule
Total training samples & 29415241 \\
Test samples (half observations) & 3455836 \\
Test samples (separate subjects) & 7221289 \\
Total test samples & 10677125 \\
\bottomrule
\end{tabularx}
\label{tab:supp-1}
\end{table}

Missing values in the dataset were handled by imputing carbohydrate intake and bolus insulin with zero. CGM measurements were linearly interpolated for gaps of up to 30 minutes, and basal or bolus insulin doses were similarly imputed with zero within the same interval. Although the set of features varies between models, identical samples were retained for both training and testing. Time-lagged features were generated over a two-hour window for the machine learning models, whereas the Loop model employed an eight-hour window to capture the complete temporal dynamics of insulin and carbohydrate effects. Lagged features were computed independently for each subject to prevent data leakage across individuals. Machine learning models incorporated CGM, insulin, and carbohydrate features, while the Loop model required separate bolus and basal insulin inputs. Any rows with remaining missing values were excluded. Feature and target values were standardized using z-score normalization for the Naïve AR and SVR models.

\subsection{Model Evaluation}

The benchmark focuses on a 30-minute prediction horizon, which is commonly used in several blood glucose prediction studies \cite{XieWang2020BenchmarkingMLBGPrediction, CapponPrendinFacchinetti2023IndividualizedModelsGlucosePrediction, LiLiuZhu2020GluNet, HameedKleinberg2020ComparingMLBGForecasting}. As suggested by Wolff et al., it is important to include metrics that capture multiple aspects of model performance \cite{WolffSchaathunGros2025BeyondAccuracyCriteria}. Accordingly, we include Root Mean Squared Error (RMSE) to measure predictive accuracy, Temporal Gain (TG) to assess the timing of predictions and the actionable lead time for interventions, and Geometric Mean (GM) to evaluate the model’s ability to detect hypo- and hyperglycemic events, accounting for class imbalance in the dataset. Although the model is trained to produce continuous glucose predictions, the classification-based GM metric is used to evaluate clinically relevant performance. In practice, clinical decisions are triggered by whether predicted glucose values cross hypo- or hyperglycemic thresholds. Training a regression model preserves granular trajectory, while evaluating predictions with GM quantifies the model’s ability to anticipate clinically critical events under class imbalance without sacrificing predictive detail \cite{AielloToffaninRiddell2025HierarchicalNetworkEnergyExpenditure}. 

All models are implemented and evaluated using GluPredKit \cite{WolffRoystonVolden2024GluPredKit}.

RMSE is defined as:
\[
\text{RMSE} = \sqrt{\frac{1}{n} \sum_{i=1}^{n} \left(Y_i - \hat{Y}_i\right)^2}
\]

where $\hat{Y}_i \in \mathbb{R}^n$ is the vector of the predicted glucose value for each data point, $Y_i \in \mathbb{R}^n$ is the vector of the corresponding actual glucose value, and n is the total number of data points. RMSE quantifies the average magnitude of prediction errors, with larger deviations penalized more heavily due to the squared differences.

TG is calculated as:
\[
TG = PH - \text{delay} \cdot T_s
\]
where
\[
\text{delay} = \arg\max_{\tau} \, E\big(y(k+\tau), \hat{y}(k)\big), \forall \, \tau \in \frac{PH}{T_s}
\]

where $y$ and $\hat{y}$ are the measured and predicted glucose values at times $k+\tau$, and $k$, respectively, $PH$ is the vector of prediction horizon instants in minutes, with PH=[0, 5, 10, 15, 20, 25, 30] minutes, and $T_S$ is the time step size equal to 5 minutes. The delay function identifies the time shift  that maximizes the estimate of the cross-correlation between the predicted and measured signals.

Temporal Gain (TG) provides insight into how accurately the model predicts the timing of key events, such as peaks and troughs in blood glucose, rather than focusing solely on overall prediction accuracy. TG values range from 0 to the prediction horizon in minutes, with the optimal value equal to the prediction horizon.

The Geometric Mean (GM) treats predictions as a classification problem for hypo- and hyperglycemia detection. Measured and predicted values are categorized into three classes: category 1 for hypoglycemia (<70 mg/dL), category 2 for in-range (70–180 mg/dL), and category 3 for hyperglycemia (>180 mg/dL). For each category $c$, the recall ($R_C$) is computed:

\[
R_C=\frac{TP_C}{TP_C+FN_C}
\]
$TP_C$ is the number of true positives for category $c$, while $FN_C$ is the number of false negatives for category $c$. The GM is then calculated as the geometric mean of the recalls across all categories:
\[
GM=\sqrt[3]{\prod_{c \in \{0,1,2\}} R_C}
\]
This metric reflects the quality of predictions within each glycemic region while accounting for class imbalance and is a useful tool to determine if a predictor is only doing well in a single type of common situation.

\subsection{Results}

The final benchmark results are presented in Table \ref{tab:supp2}. In terms of RMSE, the Naïve AR and SVR models achieve the lowest prediction errors. Conversely, the Naïve LE and PBM models perform best on GM and TG, reflecting superior timeliness of predictions and balanced detection of hypo- and hyperglycemic events. These results indicate that no single model dominates across all evaluation dimensions. Overall, each model exhibits distinct strengths, emphasizing the importance of evaluating multiple metrics when comparing predictive performance in glucose forecasting.

\begin{table}[ht!]
\centering
\caption[30-minute prediction horizon model performance]{Comparison on the testing dataset of 30-minute prediction horizon model performance using Root Mean Squared Error (RMSE – lower is better), Temporal Gain (TG – higher is better), and Geometric Mean (GM). RMSE has an optimal value of 0. TG ranges from 0 to the prediction horizon (in minutes), where the optimal value equals the prediction horizon. GM ranges from 0 to 1, with 1 being optimal.}
\begin{tabularx}{0.85\textwidth}{X S[table-format=2.0] S[table-format=2.0] S[table-format=1.2]}
\toprule
\textbf{Model} & \textbf{RMSE [mg/dL]} & \textbf{TG [mins]} & \textbf{GM} \\
\midrule
ZOH & 25 & 0 & 0.70 \\
Naïve LE & 32 & 15 & 0.83 \\
Naïve AR & 23 & 10 & 0.72 \\
PBD & 26 & 10 & 0.82 \\
Single Linear Layer NN & 24 & 5 & 0.73 \\
SVR & 23 & 10 & 0.77 \\
MLP & 24 & 5 & 0.72 \\
\bottomrule
\end{tabularx}
\label{tab:supp2}
\end{table}

As a complementary evaluation, we have included Clarke Error Grid (CEG) results in Table \ref{tab:supp3}. The CEG is a tool used to evaluate the clinical accuracy of blood glucose measurements \cite{ClarkeCoxGonderFrederick1987SMBGAccuracy}, and is also commonly used to evaluate glucose predictions. It categorizes predicted values relative to reference measurements into zones A–E, where A denotes clinically accurate predictions, B denotes benign errors that would not lead to inappropriate treatment, and C–E denote progressively more severe errors that could affect clinical decisions. The SVR has the most predictions in the A region, whereas the PBD has the fewest in the clinically unsafe regions C, D, and E, indicating lower predictive accuracy than the SVR but still a high number of predictions in the clinically benign regions.

\begin{table}[ht!]
\centering
\caption[Clarke Error Grid Analysis]{Clarke Error Grid Analysis (\%) of 30-minute ahead predictive performance for each model. Columns A–E indicate the percentage of predictions falling into each error zone (A: clinically safe, B: benign errors, C–E: increasingly severe errors).}
\begin{tabularx}{0.85\textwidth}{X S[table-format=2.1] S[table-format=2.1] S[table-format=2.1] S[table-format=2.1] S[table-format=2.1]}
\toprule
\textbf{Model} & \textbf{A [\%]} & \textbf{B [\%]} & \textbf{C [\%]} & \textbf{D [\%]} & \textbf{E [\%]} \\
\midrule
ZOH & 82.0 & 16.5 & 0.1 & 1.4 & 0.0 \\
Naïve LE & 79.0 & 19.7 & 0.6 & 0.6 & 0.0 \\
Naïve AR & 85.8 & 12.8 & 0.1 & 1.3 & 0.0 \\
PBD & 81.7 & 17.3 & 0.3 & 0.7 & 0.0 \\
Single Linear Layer NN & 83.6 & 14.9 & 0.1 & 1.3 & 0.0 \\
SVR & 86.0 & 12.9 & 0.1 & 1.0 & 0.0 \\
MLP & 83.8 & 14.8 & 0.1 & 1.3 & 0.0 \\
\bottomrule
\end{tabularx}
\label{tab:supp3}
\end{table}

\begin{figure}[ht!]
    \centering
    \includegraphics[width=1\linewidth]{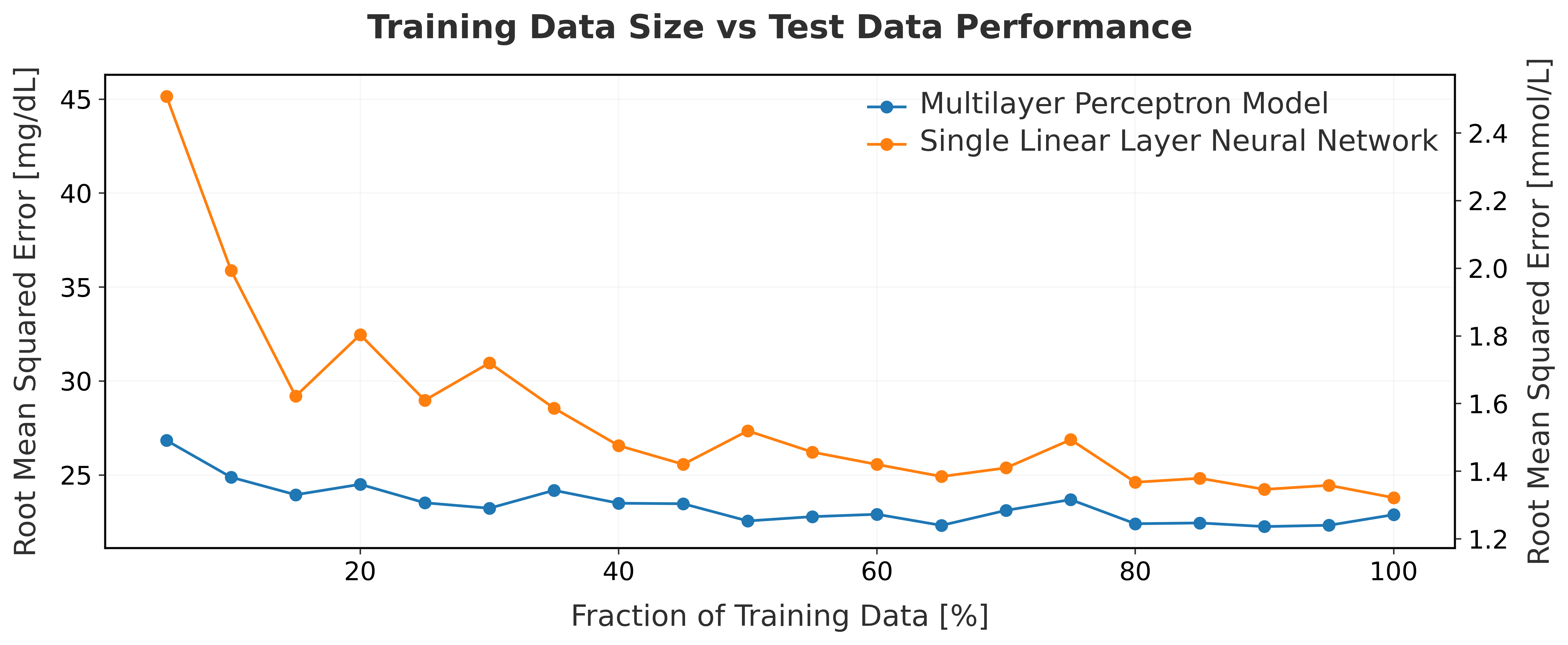}
    \caption{Comparison of a single linear layer neural network and a nonlinear multilayer perceptron (MLP) blood glucose prediction model. Test-set RMSE is plotted on the y-axis versus the percentage of available training data used for model fitting on the x-axis.}
    \label{fig:S10}
\end{figure}

A relevant point to be highlighted from the benchmark analysis is that increasing the amount of training data improves prediction accuracy on the testing data (Figure \ref{fig:S10}). Specifically, higher fractions of the available dataset lead to a reduction in RMSE in testing, illustrating the benefit of a data-rich resource like MetaboNet for model training.

\printbibliography

@article{AielloDeshpandeOzaslan2021AIDReview,
  author  = {Aiello, E. M. and Deshpande, S. and Ozaslan, B. and others},
  title   = {Review of Automated Insulin Delivery Systems for Individuals with Type 1 Diabetes: Tailored Solutions for Subpopulations},
  journal = {Current Opinion in Biomedical Engineering},
  year    = {2021},
  volume  = {19},
  pages   = {100312},
  doi     = {10.1016/j.cobme.2021.100312}
}

@article{LaffelSherrLiu2025LimitationsCGMTargets,
  author  = {Laffel, L. M. and Sherr, J. L. and Liu, J. and others},
  title   = {Limitations in Achieving Glycemic Targets From CGM Data and Persistence of Severe Hypoglycemia in Adults With Type 1 Diabetes Regardless of Insulin Delivery Method},
  journal = {Diabetes Care},
  year    = {2025},
  volume  = {48},
  number  = {2},
  pages   = {273--278},
  doi     = {10.2337/dc24-1474}
}

@article{KovatchevFrasquetPryor2024NeuralNetArtificialPancreas,
  author  = {Kovatchev, B. P. and Frasquet, A. C. and Pryor, E. C. and others},
  title   = {Neural-Net Artificial Pancreas: A Randomized Crossover Trial of a First-in-Class Automated Insulin Delivery Algorithm},
  journal = {Diabetes Technology \& Therapeutics},
  year    = {2024},
  doi     = {10.1089/dia.2023.0469}
}

@article{AielloJaloliCescon2024MPCArtificialPancreas,
  author  = {Aiello, E. M. and Jaloli, M. and Cescon, M.},
  title   = {Model Predictive Control (MPC) of an artificial pancreas with data-driven learning of multi-step-ahead blood glucose predictors},
  journal = {Control Engineering Practice},
  year    = {2024},
  volume  = {144},
  pages   = {105810},
  doi     = {10.1016/j.conengprac.2023.105810}
}

@article{matabuena2025glucodensity,
  author  = {María Matabuena and Ritobrata Ghosal and José E. Aguilar and others},
  title   = {Glucodensity Functional Profiles Outperform Traditional Continuous Glucose Monitoring Metrics},
  journal = {Scientific Reports},
  year    = {2025},
  volume  = {15},
  number  = {1},
  pages   = {33662},
  publisher = {Nature Publishing Group},
  doi     = {10.1038/s41598-025-18119-2}
}

@article{TurksoySamadiFeng2016MealDetectionModule,
  author  = {Turksoy, K. and Samadi, S. and Feng, J. and others},
  title   = {Meal Detection in Patients With Type 1 Diabetes: A New Module for the Multivariable Adaptive Artificial Pancreas Control System},
  journal = {IEEE Journal of Biomedical and Health Informatics},
  year    = {2016},
  volume  = {20},
  number  = {1},
  pages   = {47--54},
  doi     = {10.1109/JBHI.2015.2446413}
}

@article{matabuena2026functional,
  author  = {María Matabuena and Jorge Sartini and Francisco Gude},
  title   = {Beyond Scalar Metrics: Functional Data Analysis of Postprandial Continuous Glucose Monitoring in the AEGIS Study},
  journal = {BMC Medical Research Methodology},
  year    = {2026},
  volume  = {26},
  pages   = {39},
  publisher = {BioMed Central},
  doi     = {10.1186/s12874-025-02748-2}
}

@article{TurksoyBayrakQuinn2013HypoglycemiaEarlyAlarm,
  author  = {Turksoy, K. and Bayrak, E. S. and Quinn, L. and others},
  title   = {Hypoglycemia Early Alarm Systems Based On Multivariable Models},
  journal = {Industrial \& Engineering Chemistry Research},
  year    = {2013},
  volume  = {52},
  number  = {35},
  pages   = {12329--12336},
  doi     = {10.1021/ie3034015}
}

@article{AlsuhaymiBilalGarcia2025LongitudinalMultimodalDataset,
  author  = {Alsuhaymi, A. and Bilal, A. and Garc{\'i}a, D. G. and others},
  title   = {A Longitudinal Multimodal Dataset of Type 1 Diabetes},
  journal = {Scientific Data},
  year    = {2025},
  volume  = {12},
  number  = {1},
  pages   = {1379},
  doi     = {10.1038/s41597-025-05695-1}
}

@article{RiddellLiGal2023T1DEXI,
  author  = {Riddell, M. C. and Li, Z. and Gal, R. L. and others},
  title   = {Examining the Acute Glycemic Effects of Different Types of Structured Exercise Sessions in Type 1 Diabetes in a Real-World Setting: The Type 1 Diabetes and Exercise Initiative (T1DEXI)},
  journal = {Diabetes Care},
  year    = {2023},
  volume  = {46},
  number  = {4},
  pages   = {704},
  doi     = {10.2337/dc22-1721}
}

@inproceedings{MarlingBunescu2020OhioT1DMUpdate,
  author    = {Marling, C. and Bunescu, R.},
  title     = {The OhioT1DM Dataset for Blood Glucose Level Prediction: Update 2020},
  booktitle = {CEUR Workshop Proceedings},
  year      = {2020},
  volume    = {2675},
  pages     = {71--74}
}

@misc{James2025BrisT1DOpenDataset,
  author = {James, Sam Gordon},
  title  = {BrisT1D-Open Dataset},
  year   = {2025},
  doi    = {10.5523/bris.33z5jc8fa6tob21ptrugzqog08}
}

@article{PrioleauBartolomeComi2023DiaTrend,
  author  = {Prioleau, T. and Bartolome, A. and Comi, R. and others},
  title   = {DiaTrend: A dataset from advanced diabetes technology to enable development of novel analytic solutions},
  journal = {Scientific Data},
  year    = {2023},
  volume  = {10},
  number  = {1},
  pages   = {556},
  doi     = {10.1038/s41597-023-02469-5}
}

@misc{KhamesianArefeenThompson2025AZT1D,
  author = {Khamesian, S. and Arefeen, A. and Thompson, B. M. and others},
  title  = {AZT1D: A Real-World Dataset for Type 1 Diabetes},
  year   = {2025},
  doi    = {10.17632/gk9m674wcx.1},
  note   = {Version 1}
}

@article{MaheshwariKaliaTewari2025AIForDiabetesReview,
  author  = {Maheshwari, S. and Kalia, A. and Tewari, J. and others},
  title   = {Artificial intelligence for diabetes management -- a review},
  journal = {JDMDC},
  year    = {2025},
  volume  = {12},
  number  = {1},
  pages   = {24--32},
  doi     = {10.15406/jdmdc.2025.12.00292}
}

@inproceedings{DengDongSocher2009ImageNet,
  author    = {Deng, J. and Dong, W. and Socher, R. and others},
  title     = {ImageNet: A Large-Scale Hierarchical Image Database},
  booktitle = {Proceedings of the IEEE Conference on Computer Vision and Pattern Recognition (CVPR)},
  year      = {2009},
  doi       = {10.1109/CVPR.2009.5206848}
}

@article{ChawlaNakovAli2023TenYearsAfterImageNet,
  author  = {Chawla, S. and Nakov, P. and Ali, A. and others},
  title   = {Ten years after ImageNet: a 360\textdegree\ perspective on artificial intelligence},
  journal = {Royal Society Open Science},
  year    = {2023},
  volume  = {10},
  number  = {3},
  pages   = {221414},
  doi     = {10.1098/rsos.221414}
}

@online{AnonymousNDDPublicStudyWebsitesJAEB,
  author = {JAEB Center for Health Research},
  title  = {Diabetes Datasets and Documents},
  url    = {https://public.jaeb.org/datasets/diabetes},
  urldate   = {2025-10-26}
}

@online{AnonymousNDDBabelbetes,
  author = {{Nudge BG}},
  title  = {Babelbetes},
  url    = {https://nudgebg.github.io/babelbetes/},
  urldate   = {2025-10-26}
}

@article{DelGiudicePiersantiGobl2025OpenDynamicGlycemicData,
  author  = {Del Giudice, L. L. and Piersanti, A. and G{\"o}bl, C. and others},
  title   = {Availability of Open Dynamic Glycemic Data in the Field of Diabetes Research: A Scoping Review},
  journal = {Journal of Diabetes Science and Technology},
  year    = {2025},
  pages   = {19322968251316896},
  doi     = {10.1177/19322968251316896}
}

@article{ShahidLewis2022LargeScaleDataAnalysisGlucoseVariability,
  author  = {Shahid, A. and Lewis, D. M.},
  title   = {Large-Scale Data Analysis for Glucose Variability Outcomes with Open-Source Automated Insulin Delivery Systems},
  journal = {Nutrients},
  year    = {2022},
  volume  = {14},
  number  = {9},
  pages   = {1906},
  doi     = {10.3390/nu14091906}
}

@article{NeinsteinWongLook2016TidepoolPlatform,
  author  = {Neinstein, A. and Wong, J. and Look, H. and others},
  title   = {A case study in open source innovation: developing the Tidepool Platform for interoperability in type 1 diabetes management},
  journal = {Journal of the American Medical Informatics Association},
  year    = {2016},
  volume  = {23},
  number  = {2},
  pages   = {324--332},
  doi     = {10.1093/jamia/ocv104}
}

@misc{ReplicahealthNDMetaboNetGitHub,
  author = {{Replica Health}},
  title  = {MetaboNet Processor},
  year   = {n.d.},
  url    = {https://github.com/replicahealth/metabonet_processor},
  note   = {Accessed 2025-10-26}
}

@article{HidalgoAlvaradoBotella2024HUPAUCM,
  author  = {Hidalgo, J. I. and Alvarado, J. and Botella, M. and others},
  title   = {HUPA-UCM diabetes dataset},
  journal = {Data in Brief},
  year    = {2024},
  volume  = {55},
  pages   = {110559},
  doi     = {10.1016/j.dib.2024.110559}
}

@article{cinar2025benchmarking,
  author  = {Beyza Cinar and Maria Maleshkova},
  title   = {Benchmarking Hypoglycemia Classification Using Quality-Enhanced DiaData},
  journal = {IEEE Journal of Biomedical and Health Informatics},
  year    = {2025},
  volume  = {29},
  number  = {12},
  pages   = {8831--8838},
  month   = dec,
  publisher = {IEEE},
  doi     = {10.1109/JBHI.2025.3620603}
}

@inproceedings{cinar2025diadata,
  author    = {Beyza Cinar and Maria Maleshkova},
  title     = {DiaData: An Integrated Large Dataset for Type 1 Diabetes and Hypoglycemia Research},
  booktitle = {BIO Web of Conferences},
  year      = {2025},
  volume    = {195},
  pages     = {03001},
  editor    = {R. Hofestädt},
  publisher = {EDP Sciences},
  doi       = {10.1051/bioconf/202519503001}
}

@article{rand2026cgmreview,
  author  = {E. Jr and Christopher Rand and A. Ayers and others},
  title   = {A Comprehensive Review of the Evolving Landscape of Continuous Glucose Monitoring Metrics},
  journal = {JMIR Preprints},
  year    = {2026},
  note    = {Preprint},
  doi     = {10.2196/preprints.92455}
}

@article{bergford2023exercise,
  author  = {S. Bergford and Michael C. Riddell and Peter G. Jacobs and others},
  title   = {The Type 1 Diabetes and Exercise Initiative: Predicting Hypoglycemia Risk During Exercise for Participants with Type 1 Diabetes Using Repeated Measures Random Forest},
  journal = {Diabetes Technology \& Therapeutics},
  year    = {2023},
  volume  = {25},
  number  = {9},
  pages   = {602--611},
  publisher = {Mary Ann Liebert},
  doi     = {10.1089/dia.2023.0140}
}

@article{leutheuser2024nocturnal,
  author  = {H. Leutheuser and M. Bartholet and A. Marx and others},
  title   = {Predicting Risk for Nocturnal Hypoglycemia After Physical Activity in Children with Type 1 Diabetes},
  journal = {Frontiers in Medicine},
  year    = {2024},
  volume  = {11},
  pages   = {1439218},
  publisher = {Frontiers Media},
  doi     = {10.3389/fmed.2024.1439218}
}

@inproceedings{piersanti2023ml,
  author    = {A. Piersanti and B. Salvatori and C. G{\"o}bl and others},
  title     = {A Machine-Learning Framework Based on Continuous Glucose Monitoring to Prevent the Occurrence of Exercise-Induced Hypoglycemia in Children with Type 1 Diabetes},
  booktitle = {Proceedings of the 2023 IEEE 36th International Symposium on Computer-Based Medical Systems (CBMS)},
  year      = {2023},
  pages     = {281--286},
  publisher = {IEEE},
  doi       = {10.1109/CBMS58004.2023.00231}
}

@article{calzavara2025systematic,
  author  = {Andrea Calzavara and Filippo Prendin and Giovanni Cappon and others},
  title   = {Systematic Review on Deep Learning Algorithms for Blood Glucose Forecasting in Type 1 Diabetes},
  journal = {IEEE Journal of Biomedical and Health Informatics},
  year    = {2025},
  note    = {Early access (in press)},
  publisher = {IEEE},
  doi     = {10.1109/JBHI.2025.3630214}
}

@techreport{grieco2001race,
  author      = {Elizabeth M. Grieco and Rachel C. Cassidy},
  title       = {Overview of Race and Hispanic Origin: 2000},
  institution = {U.S. Census Bureau},
  year        = {2001},
  month       = apr,
  number      = {C2KBR/01-1},
  type        = {Census 2000 Brief},
  url         = {https://www.census.gov/library/publications/2001/dec/c2kbr01-01.html},
  note        = {Accessed: 2026-02-25}
}

@article{ZhaoZhuShen2023ChineseDiabetesDatasets,
  author  = {Zhao, Q. and Zhu, J. and Shen, X. and others},
  title   = {Chinese diabetes datasets for data-driven machine learning},
  journal = {Scientific Data},
  year    = {2023},
  volume  = {10},
  number  = {1},
  pages   = {35},
  doi     = {10.1038/s41597-023-01940-7}
}

@article{WolffRoystonFougner2025HarmonizingDiabetesDatasets,
  author  = {Wolff, M. K. and Royston, S. and Fougner, A. L. and others},
  title   = {A perspective on harmonizing diabetes management datasets},
  journal = {Data in Brief},
  year    = {2025},
  volume  = {59},
  pages   = {111399},
  doi     = {10.1016/j.dib.2025.111399}
}

@online{jaeb_ctr3_581,
  author       = {{JAEB Center for Health Research}},
  title        = {Pilot Study 3 of Outpatient Control-to-Range: Safety and Efficacy with Day-and-Night In-Home Use (CTR3)},
  year         = {n.d.},
  url          = {https://public.jaeb.org/dataset/581},
  urldate      = {2026-01-12}
}

@online{jaeb_idcl_dclp3_573,
  author       = {{JAEB Center for Health Research}},
  title        = {The International Diabetes Closed Loop (iDCL) Trial: Clinical Acceptance of the Artificial Pancreas - A Pivotal Study of t:Slim X2 with Control-IQ Technology (DCLP3)},
  year         = {n.d.},
  url          = {https://public.jaeb.org/dataset/573},
  urldate      = {2026-01-12}
}

@online{jaeb_dclp5_535,
  author       = {{JAEB Center for Health Research}},
  title        = {A Multi-Center Study of the Control-IQ Closed Loop Control System in Children with Type 1 Diabetes (DCLP5)},
  year         = {n.d.},
  url          = {https://public.jaeb.org/dataset/535},
  urldate      = {2026-01-12}
}

@online{jaeb_flair_566,
  author       = {{JAEB Center for Health Research}},
  title        = {FLAIR — Fuzzy Logic Automated Insulin Regulation: A Crossover Study Comparing Two Automated Insulin Delivery System Algorithms (PID vs.\ PID + Fuzzy Logic) in Individuals with Type 1 Diabetes},
  year         = {n.d.},
  url          = {https://public.jaeb.org/dataset/566},
  urldate      = {2026-01-12}
}

@online{jaeb_ilet_579_iobp2,
  author       = {{JAEB Center for Health Research}},
  title        = {The Insulin-Only Bionic Pancreas Pivotal Trial: Testing the iLet in Adults and Children with Type 1 Diabetes},
  year         = {n.d.},
  url          = {https://public.jaeb.org/dataset/579},
  urldate      = {2026-01-12}
}

@online{jaeb_loop_560,
  author       = {{JAEB Center for Health Research}},
  title        = {An Observational Study of Individuals with Type 1 Diabetes Using the Loop System for Automated Insulin Delivery},
  year         = {n.d.},
  url          = {https://public.jaeb.org/dataset/560},
  urldate      = {2026-01-12}
}

@online{jaeb_pedap_599,
  author       = {{JAEB Center for Health Research}},
  title        = {The Pediatric Artificial Pancreas (PEDAP) Trial of Control-IQ Technology in Young Children in Type 1 Diabetes},
  year         = {n.d.},
  url          = {https://public.jaeb.org/dataset/599},
  urldate      = {2026-01-12}
}

@online{jaeb_cgm_546_replacebg,
  author       = {{JAEB Center for Health Research}},
  title        = {A Randomized Trial Comparing Continuous Glucose Monitoring With and Without Routine Blood Glucose Monitoring in Adults with Type 1 Diabetes},
  year         = {n.d.},
  url          = {https://public.jaeb.org/dataset/546},
  urldate      = {2026-01-12}
}

@misc{MetaboNetNDDataDictionary,
  author = {{Replica Health}},
  title  = {MetaboNet},
  year   = {n.d.},
  url    = {https://metabo-net.org/data-dictionary},
  note   = {Accessed 2025-11-07}
}

@article{CooperReinholdShahid2025GlucoseVariabilityTwoDatasets,
  author  = {Cooper, D. and Reinhold, B. and Shahid, A. and others},
  title   = {Glucose Variability Analysis in Two Large-Scale and Real-World Data Sets of Open-Source Automated Insulin Delivery Systems},
  journal = {Journal of Diabetes Science and Technology},
  year    = {2025},
  volume  = {19},
  number  = {3},
  pages   = {649--657},
  doi     = {10.1177/19322968231198871}
}

@article{ChoAielloOzaslan2024PhysicalActivityDetectionFramework,
  author  = {Cho, S. and Aiello, E. M. and Ozaslan, B. and others},
  title   = {Design of a Real-Time Physical Activity Detection and Classification Framework for Individuals With Type 1 Diabetes},
  journal = {Journal of Diabetes Science and Technology},
  year    = {2024},
  volume  = {18},
  number  = {5},
  pages   = {1146--1156},
  doi     = {10.1177/19322968231153896}
}

@article{DhaliwalTangAiello2025HypoglycemiaRiskPhysicalActivity,
  author  = {Dhaliwal, M. and Tang, K. and Aiello, E. M. and others},
  title   = {Variation in Hypoglycemia Risk During Real-World Physical Activity in Adults with Type 1 Diabetes: Insights from the Type 1 Diabetes Exercise Initiative},
  journal = {Diabetes Technology \& Therapeutics},
  year    = {2025},
  pages   = {15209156251400209},
  doi     = {10.1177/15209156251400209}
}

@article{PionaAielloMancioppi2025PrediabetesYouthObesity,
  author  = {Piona, C. and Aiello, E. M. and Mancioppi, V. and others},
  title   = {An Exploratory Analysis of Continuous Glucose Monitoring Metrics in Relation to Prediabetes in Youths with Obesity},
  journal = {Diabetes Technology \& Therapeutics},
  year    = {2025},
  doi     = {10.1177/15209156251407959}
}

@article{DunnAjjanBergenstal2024BeyondTIRtoTITR,
  author  = {Dunn, T. C. and Ajjan, R. A. and Bergenstal, R. M. and others},
  title   = {Is It Time to Move Beyond TIR to TITR? Real-World Data from Over 20,000 Users of Continuous Glucose Monitoring in Patients with Type 1 and Type 2 Diabetes},
  journal = {Diabetes Technology \& Therapeutics},
  year    = {2024},
  volume  = {26},
  number  = {3},
  pages   = {203--210},
  doi     = {10.1089/dia.2023.0565}
}

@article{BeckRaghinaruCalhoun2024TIRvsTITR,
  author  = {Beck, R. W. and Raghinaru, D. and Calhoun, P. and others},
  title   = {A Comparison of Continuous Glucose Monitoring-Measured Time-in-Range 70-180 mg/dL Versus Time-in-Tight-Range 70-140 mg/dL},
  journal = {Diabetes Technology \& Therapeutics},
  year    = {2024},
  volume  = {26},
  number  = {3},
  pages   = {151--155},
  doi     = {10.1089/dia.2023.0380}
}

@misc{FuNorouziNachum2021BenchmarksDeepOPE,
  author       = {Fu, J. and Norouzi, M. and Nachum, O. and others},
  title        = {Benchmarks for Deep Off-Policy Evaluation},
  year         = {2021},
  doi          = {10.48550/arXiv.2103.16596},
  archivePrefix= {arXiv},
  eprint       = {2103.16596}
}

@article{EmersonGuyMcConville2023OfflineRLSaferBG,
  author  = {Emerson, H. and Guy, M. and McConville, R.},
  title   = {Offline reinforcement learning for safer blood glucose control in people with type 1 diabetes},
  journal = {Journal of Biomedical Informatics},
  year    = {2023},
  volume  = {142},
  pages   = {104376},
  doi     = {10.1016/j.jbi.2023.104376}
}

@article{ZhuLiGeorgiou2023OfflineDRLBasalInsulin,
  author  = {Zhu, T. and Li, K. and Georgiou, P.},
  title   = {Offline Deep Reinforcement Learning and Off-Policy Evaluation for Personalized Basal Insulin Control in Type 1 Diabetes},
  journal = {IEEE Journal of Biomedical and Health Informatics},
  year    = {2023},
  volume  = {27},
  number  = {10},
  pages   = {5087--5098},
  doi     = {10.1109/JBHI.2023.3303367}
}

@misc{LevineKumarTucker2020OfflineRLTutorial,
  author       = {Levine, S. and Kumar, A. and Tucker, G. and others},
  title        = {Offline Reinforcement Learning: Tutorial, Review, and Perspectives on Open Problems},
  year         = {2020},
  doi          = {10.48550/arXiv.2005.01643},
  archivePrefix= {arXiv},
  eprint       = {2005.01643}
}

@article{BattelinoDanneBergenstal2019TIRConsensus,
  author  = {Battelino, Tadej and Danne, Thomas and Bergenstal, Richard M. and others},
  title   = {Clinical Targets for Continuous Glucose Monitoring Data Interpretation: Recommendations From the International Consensus on Time in Range},
  journal = {Diabetes Care},
  year    = {2019},
  volume  = {42},
  number  = {8},
  pages   = {1593--1603},
  doi     = {10.2337/dci19-0028}
}

@book{HastieTibshiraniFriedman2009ESL,
  author    = {Hastie, Trevor and Tibshirani, Robert and Friedman, Jerome},
  title     = {The Elements of Statistical Learning},
  series    = {Springer Series in Statistics},
  publisher = {Springer},
  address   = {New York, NY},
  year      = {2009},
  doi       = {10.1007/978-0-387-84858-7}
}

@article{JacobsHerreroFacchinetti2023AIMLBestPractices,
  author  = {Jacobs, Peter G. and Herrero, Pau and Facchinetti, Andrea and others},
  title   = {Artificial intelligence and machine learning for improving glycemic control in diabetes: best practices, pitfalls and opportunities},
  journal = {IEEE Reviews in Biomedical Engineering},
  year    = {2023},
  note    = {Early Access},
  doi     = {10.1109/RBME.2023.3331297}
}

@article{ChawlaMakkar2019RSSDIInsulinConsensus,
  author  = {{Expert panel (extended)} and Chawla, R. and Makkar, B. M. and others},
  title   = {RSSDI consensus recommendations on insulin therapy in the management of diabetes},
  journal = {International Journal of Diabetes in Developing Countries},
  year    = {2019},
  volume  = {39},
  number  = {S2},
  pages   = {43--92},
  doi     = {10.1007/s13410-019-00783-6}
}

@article{ChakrabartyZavitsanouDoyle2018EventTriggeredMPC,
  author  = {Chakrabarty, Ankush and Zavitsanou, Styliani and Doyle, Francis J. and others},
  title   = {Event-Triggered Model Predictive Control For Embedded Artificial Pancreas Systems},
  journal = {IEEE Transactions on Biomedical Engineering},
  year    = {2018},
  volume  = {65},
  number  = {3},
  pages   = {575--586},
  doi     = {10.1109/TBME.2017.2707344}
}

@article{vanHeusdenDassauZisser2012ControlRelevantModels,
  author  = {van Heusden, K. and Dassau, E. and Zisser, H. C. and others},
  title   = {Control-Relevant Models for Glucose Control Using A Priori Patient Characteristics},
  journal = {IEEE Transactions on Biomedical Engineering},
  year    = {2012},
  volume  = {59},
  number  = {7},
  pages   = {1839--1849},
  doi     = {10.1109/TBME.2011.2176939}
}

@article{XieWang2020BenchmarkingMLBGPrediction,
  author  = {Xie, J. and Wang, Q.},
  title   = {Benchmarking Machine Learning Algorithms on Blood Glucose Prediction for Type I Diabetes in Comparison With Classical Time-Series Models},
  journal = {IEEE Transactions on Biomedical Engineering},
  year    = {2020},
  volume  = {67},
  number  = {11},
  pages   = {3101--3124},
  doi     = {10.1109/TBME.2020.2975959}
}

@article{CapponPrendinFacchinetti2023IndividualizedModelsGlucosePrediction,
  author  = {Cappon, Giulia and Prendin, Francesco and Facchinetti, Andrea and others},
  title   = {Individualized Models for Glucose Prediction in Type 1 Diabetes: Comparing Black-Box Approaches to a Physiological White-Box One},
  journal = {IEEE Transactions on Biomedical Engineering},
  year    = {2023},
  volume  = {70},
  number  = {11},
  pages   = {3105--3115},
  doi     = {10.1109/TBME.2023.3276193}
}

@article{LiLiuZhu2020GluNet,
  author  = {Li, K. and Liu, C. and Zhu, T. and others},
  title   = {GluNet: A Deep Learning Framework for Accurate Glucose Forecasting},
  journal = {IEEE Journal of Biomedical and Health Informatics},
  year    = {2020},
  volume  = {24},
  number  = {2},
  pages   = {414--423},
  doi     = {10.1109/JBHI.2019.2931842}
}

@inproceedings{HameedKleinberg2020ComparingMLBGForecasting,
  author    = {Hameed, H. and Kleinberg, S.},
  title     = {Comparing Machine Learning Techniques for Blood Glucose Forecasting Using Free-living and Patient Generated Data},
  booktitle = {Proceedings of Machine Learning Research},
  year      = {2020},
  volume    = {126},
  pages     = {871--894}
}

@article{WolffSchaathunGros2025BeyondAccuracyCriteria,
  author  = {Wolff, M. K. and Schaathun, H. G. and Gros, S. and others},
  title   = {Blood Glucose Prediction Algorithms Require Clinically Relevant Performance Criteria Beyond Accuracy},
  journal = {Diabetes Technology \& Therapeutics},
  year    = {2025},
  volume  = {27},
  number  = {10},
  pages   = {858--870},
  doi     = {10.1089/dia.2025.0074}
}

@article{AielloToffaninRiddell2025HierarchicalNetworkEnergyExpenditure,
  author  = {Aiello, E. M. and Toffanin, C. and Riddell, M. C. and others},
  title   = {A hierarchical network model for the estimate of the energy expenditure in individuals with type 1 diabetes},
  journal = {Engineering Applications of Artificial Intelligence},
  year    = {2025},
  volume  = {159},
  pages   = {111758},
  doi     = {10.1016/j.engappai.2025.111758}
}

@article{qaraqe2024fewshot,
  author  = {Mohammad Qaraqe and Ahmad Elzein and Samir Belhaouari and others},
  title   = {A Novel Few-Shot Learning Derived Architecture for Long-Term HbA1c Prediction},
  journal = {Scientific Reports},
  year    = {2024},
  volume  = {14},
  number  = {1},
  pages   = {482},
  publisher = {Nature Publishing Group},
  doi     = {10.1038/s41598-023-50348-1}
}

@article{WolffRoystonVolden2024GluPredKit,
  author  = {Wolff, M. K. and Royston, S. and Volden, R.},
  title   = {GluPredKit: A Python Package for Blood Glucose Prediction and Evaluation},
  journal = {Journal of Open Source Software},
  year    = {2024},
  volume  = {9},
  number  = {101},
  pages   = {6904},
  doi     = {10.21105/joss.06904}
}

@article{ClarkeCoxGonderFrederick1987SMBGAccuracy,
  author  = {Clarke, W. L. and Cox, D. and Gonder-Frederick, L. A. and others},
  title   = {Evaluating clinical accuracy of systems for self-monitoring of blood glucose},
  journal = {Diabetes Care},
  year    = {1987},
  volume  = {10},
  number  = {5},
  pages   = {622--628},
  doi     = {10.2337/diacare.10.5.622}
}
\end{document}